\documentclass[letterpaper]{article} 
\usepackage{aaai25}  
\usepackage{times}  
\usepackage{helvet}  
\usepackage{courier}  
\usepackage[hyphens]{url}  
\usepackage{graphicx} 
\usepackage{natbib}  
\usepackage{caption} 
\usepackage{algorithm}
\usepackage{algorithmic}
\usepackage{amsmath}
\usepackage{amssymb}
\usepackage{multirow}
\usepackage{booktabs}

\usepackage{newfloat}
\usepackage{listings}
\DeclareCaptionStyle{ruled}{labelfont=normalfont,labelsep=colon,strut=off} 
\floatstyle{ruled}
\newfloat{listing}{tb}{lst}{}
\floatname{listing}{Listing}
\title{SpatioTemporal Learning for Human Pose Estimation in Sparsely-Labeled Videos}
\author{
    Yingying Jiao\textsuperscript{\rm 1,2}\equalcontrib, 
    Zhigang Wang\textsuperscript{\rm 3}\equalcontrib\thanks{Corresponding authors.},
    Sifan Wu\textsuperscript{\rm 1,2},
    Shaojing Fan\textsuperscript{\rm 4}\footnotemark[2],\\
    Zhenguang Liu\textsuperscript{\rm 5,6}\footnotemark[2],
    Zhuoyue Xu\textsuperscript{\rm 3},
    Zheqi Wu\textsuperscript{\rm 3}
}
\affiliations{
    \textsuperscript{\rm 1}College of Computer Science and Technology, Jilin University\\
    \textsuperscript{\rm 2}Key Laboratory of Symbolic Computation and Knowledge Engineering of Ministry of Education, Jilin University\\
    \textsuperscript{\rm 3}College of Computer Science and Technology, Zhejiang Gongshang University\\
    \textsuperscript{\rm 4}School of Computing, National University of Singapore\\
    \textsuperscript{\rm 5}The State Key Laboratory of Blockchain and Data Security, Zhejiang University\\
    \textsuperscript{\rm 6}Hangzhou High-Tech Zone (Binjiang) Institute of Blockchain and Data Security\\
    
    jiaoyy21@mails.jlu.edu.cn, wangzhigang2024@gmail.com, wusifan2021@gmail.com, fanshaojing@gmail.com,\\
    liuzhenguang2008@gmail.com, 1069516849xzyy@gmail.com, chasewoo17@gmail.com
    
}

\begin{document}

\maketitle

\begin{abstract}
Human pose estimation in videos remains a challenge, largely due to the reliance on extensive manual annotation of large datasets, which is expensive and labor-intensive. Furthermore, existing approaches often struggle to capture long-range temporal dependencies and overlook the complementary relationship between temporal pose heatmaps and visual features. To address these limitations, we introduce STDPose, a novel framework that enhances human pose estimation by learning spatiotemporal dynamics in sparsely-labeled videos. STDPose incorporates two key innovations: 1) A novel Dynamic-Aware Mask to capture long-range motion context, allowing for a nuanced understanding of pose changes. 2) A system for encoding and aggregating spatiotemporal representations and motion dynamics to effectively model spatiotemporal relationships, improving the accuracy and robustness of pose estimation. STDPose establishes a new performance benchmark for both video pose propagation (\textit{i.e.}, propagating pose annotations from labeled frames to unlabeled frames) and pose estimation tasks, across three large-scale evaluation datasets. Additionally, utilizing pseudo-labels generated by pose propagation, STDPose achieves competitive performance with only 26.7\% labeled data.
\end{abstract}

%

\section{Introduction}

In recent years, visual perception tasks~\cite{dosovitskiy2020image, kirillov2023segment, shuai2023locate} have achieved significant research breakthroughs, largely owing to the continuous advancement of model architectures~\cite{vaswani2017attention, liu2022convnet} and the rollout of large-scale datasets~\cite{deng2009imagenet, lin2014microsoft}. Correspondingly, \textit{human pose estimation}~\cite{liu2021deep, sun2019deep}, as a foundational task in computer vision~\cite{chen2023activating, wang2022internvideo}, has flourished over the past few years and is particularly valuable in a wide range of applications, including \textit{sports analytics}, \textit{surveillance}, \textit{augmented reality}, and \textit{human-computer interaction}~\cite{schmidtke2021unsupervised, tse2022collaborative, yang2023action, liu2022copy, wu2024pose, su2021motion}.

\begin{figure}[t]
\centering
\includegraphics[width=.99\linewidth]{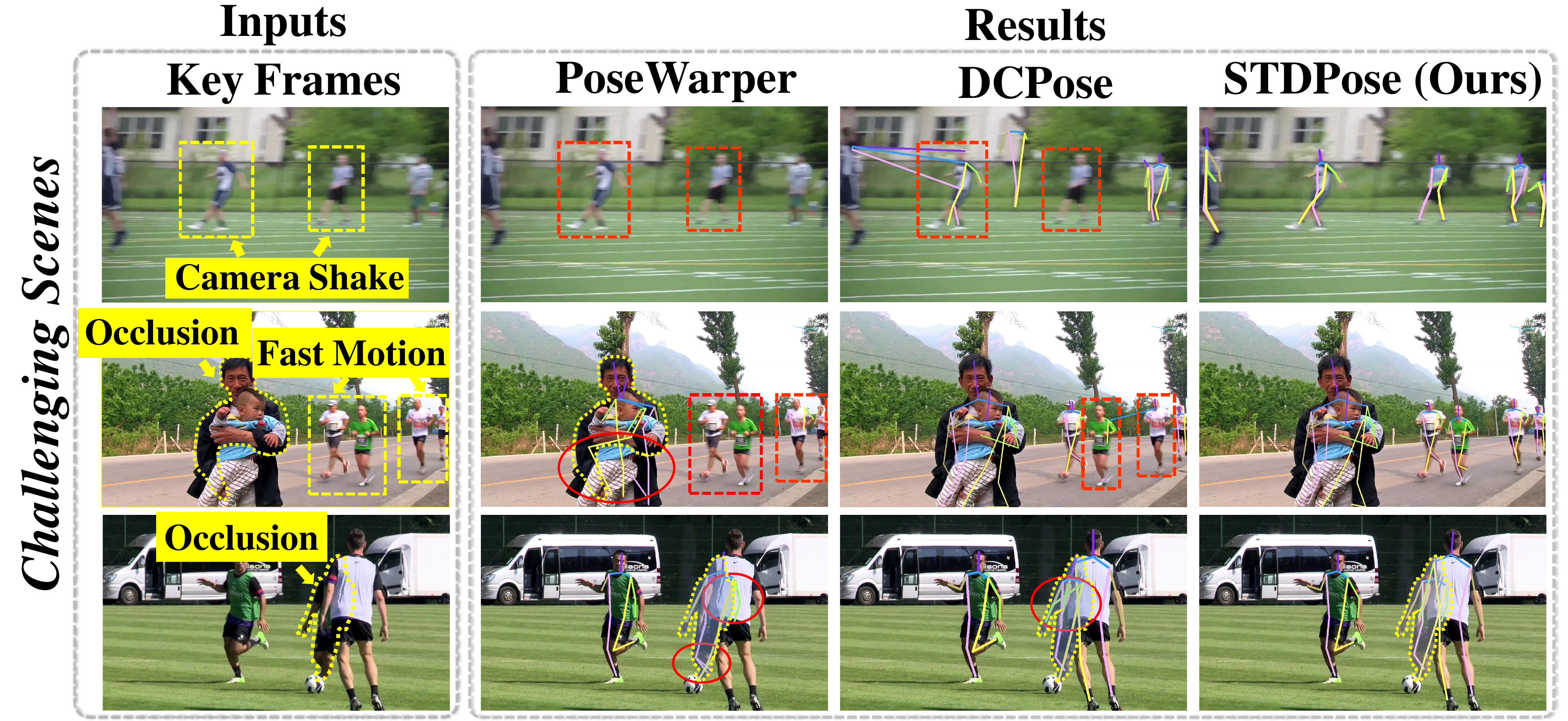}
\caption{Our model, STDPose, consistently demonstrates high accuracy and robustness in human pose estimation, even in challenging video scenes with \textbf{blur} and \textbf{occlusion}, thanks to its innovative approach to capturing spatiotemporal information and long-range motion cues. However, state-of-the-art methods like PoseWarper~\cite{bertasius2019learning} and DCPose~\cite{liu2021deep} struggle in such scenarios. Red rectangles in our visual comparisons indicate where these methods completely failed to detect \textbf{blurred} individuals, while red ellipses highlight their incorrect detections of wrist and ankle joints caused by severe \textbf{occlusion}. All results are from models trained on sparsely labeled (\textit{i.e.}, every 7 frames) videos.} 

\label{fig:comparison}
\end{figure}

However, the development of multi-person video pose estimation has plateaued, facing numerous challenges that demand innovative solutions. \textbf{First}, traditional model architectures, originally designed for static images, continue to struggle with common video issues such as pose occlusion and blur due to fast motion or camera shake. \textbf{Second}, most leading methods heavily rely on large-scale benchmark datasets that require intensive time-consuming and labor-intensive pose annotations. Notably, the temporal information in video sequences often shows substantial redundancy, and the spatial changes in pose from one frame to the next are typically minor~\cite{bertasius2019learning}, resulting in substantial repetitive effort for manual annotation.

Efforts to overcome the inherent limitations of traditional static image-based methods for human pose estimation have propelled substantial progress. Video-focused methods \cite{bertasius2019learning, wang2020combining} are designed to capture the temporal dependencies and correlations within video data, which are often neglected by conventional static image methods \cite{newell2016stacked, sun2019deep}. For instance, PoseWarper~\cite{bertasius2019learning} and DCPose~\cite{liu2021deep} model pose residuals for aggregating temporal contexts, utilizing deformable convolutions~\cite{zhu2019deformable}.

Empirically, we observe that in challenging scenes such as pose occlusion and blur, existing methods~\cite{bertasius2019learning, liu2021deep} frequently exhibit poor performance, as illustrated in Figure~\ref{fig:comparison}. Our experiments and analyses suggest multifaceted reasons for these shortcomings:  
{\textbf{(1)}} Current state-of-the-art methods directly aggregate spatiotemporal representatons, lacking mathematical constraints guaranteeing that label-relevant clues are extracted and redundant information is reduced, which leads to suboptimal pose estimation outcomes. Furthermore, in scenarios of pose occlusion and blur, temporally distant frames may carry more pertinent supplementary evidence than adjacent frames due to the temporal similarity between consecutive frames, yet current methods~\cite{liu2021deep, bertasius2019learning} fail to effectively harness these long-range spatiotemporal contexts. 
{\textbf{(2)}} Despite these approaches~\cite{liu2021deep, liu2022temporal} employing deformable convolutions to simulate various receptive fields, they excessively emphasize local pose variations and neglect global spatial correlations, potentially limiting a comprehensive understanding. {\textbf{(3)}} Recent methods~\cite{liu2022temporal, feng2023mutual, wu2024joint} focus solely on temporal features and overlook the integration of complementary information from pose heatmaps, consequently restricting the incorporation of pose annotations in pose propagation.

To address the limitations of existing methods, we present a novel framework, termed \textbf{STDPose}, which encodes and aggregates spatiotemporal representations and motion dynamics to learn \textbf{\underline{S}}patio\textbf{\underline{T}}emporal \textbf{\underline{D}}ynamics for Human \textbf{\underline{Pose}} Estimation. The motivation behind STDPose is to propagate pose annotations from labeled (auxiliary) frames to unlabeled (key) frames within sparsely-labeled videos at intervals (\textit{i.e.}, every $T$ frames), to reduce manual labor. In particular, our STDPose embraces two key components: \textbf{(i)} A SpatioTemporal Representation Encoder (STRE) is proposed to address limitations (1) and (3) by collaboratively integrating multi-frame visual features and pose heatmaps to comprehensively capture spatiotemporal dependencies through its two designed specialized submodules. Additionally, a mutual information objective is utilized to supervise cross-frame task-relevant knowledge extraction. \textbf{(ii)} To address limitations (1) and (2), we further introduce a novel Dynamic-Aware Mask (DAM) that dedicated to effectively capture long-range motion contexts through a modified sigmoid function. Finally, the SpatioTemporal Dynamics Aggregation module (STDA) aggregates spatiotemporal representations and motion dynamics, enhancing spatial and temporal coherence. 
Our method achieves state-of-the-art results for both the video pose propagation and video pose estimation tasks across three benchmark datasets. To summarize, the main contributions of this paper are as follows:

\begin{itemize}
\item We develop a novel framework that effectively learns SpatioTemporal Dynamics for Human Pose Estimation in sparsely-labeled videos by encoding and aggregating spatiotemporal representations and motion details, enhancing pose estimation accuracy and robustness through leveraging the temporal continuity between frames.

\item We introduce a pioneering technique called the Dynamic-Aware Mask (DAM), which dynamically captures long-range motion clues. This allows for a more nuanced understanding of pose offsets, especially in areas prone to occlusion or blur.

\item Our method STDPose significantly advances the field by establishing new state-of-the-art benchmarks for both the pose propagation task and standard pose estimation task across three challenging benchmark datasets: PoseTrack2017, PoseTrack2018, and PoseTrack2021. By automatically generating accurate pose annotations throughout the entire video with only a few manually labeled frames, STDPose reduces the reliance on labor-intensive manual annotation and provides the research community with new insights for video pose estimation.
\end{itemize}

\begin{figure*}[t]
\centering
\includegraphics[width=0.99\linewidth,keepaspectratio]{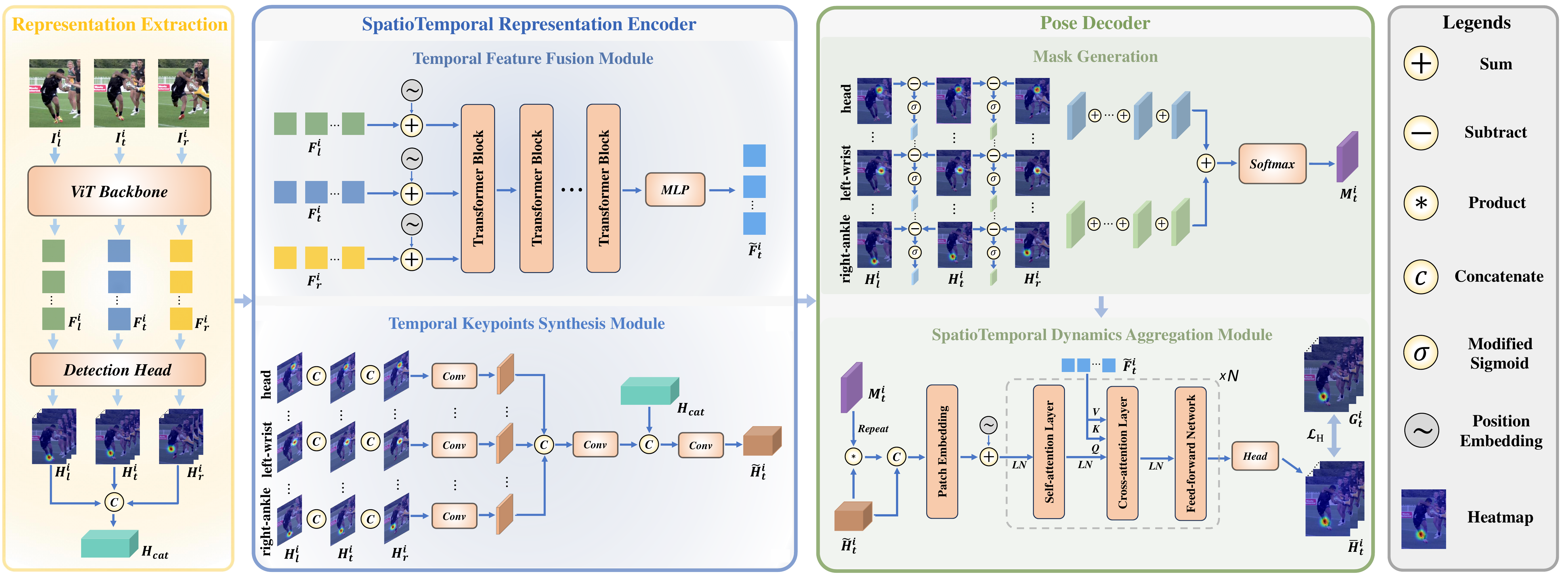}
\caption{The overall pipeline of our STDPose framework. Given an input sequence $\left\langle I_{l}^{i},  I_{t}^{i} , I_{r}^{i}\right\rangle$, our goal is to estimate the human pose of the key frame $I_{t}^{i}$.} 
\label{fig:framework}
\end{figure*}

\section{Related Work}

\textbf{Image-based human pose estimation.} Traditional human pose estimation methods rely on pictorial structures~\cite{zhang2009efficient, sapp2010cascaded} but are limited by handcrafted features and poor generalization. Inspired by advances in deep learning and benchmarks like PoseTrack~\cite{doering2022posetrack21} and COCO~\cite{lin2014microsoft}, recent deep learning based approaches~\cite{artacho2020unipose, wei2016convolutional, xiao2018simple} have emerged. HRNet~\cite{sun2019deep} excels with high-resolution features, and vision transformers~\cite{dosovitskiy2020image, xu2022vitpose} show promising performance. However, image-based methods struggle with video-specific challenges like blur and occlusion due to a lack of temporal context. This work advances by focusing on video-based pose estimation to address these dynamic issues effectively.

\textbf{Video-based human pose estimation.} To address the shortcomings of image-based approaches in videos, video-based methods~\cite{bertasius2019learning, liu2021deep, liu2022temporal, feng2023mutual, he2024video, jin2022otpose} focus on learning temporal information from neighboring frames. Techniques such as calculating dense optical flow~\cite{pfister2015flowing, ilg2017flownet, xiu2018pose} between frames use enriched flow representations to enhance prediction accuracy. DetTrack~\cite{wang2020combining} integrates 3D-HRNet and 3D convolutions for slight improvements in pose sequence estimation. However, existing methods struggle with three main issues: i) dependence on extensively labeled data; ii) difficulty capturing long-range temporal dependencies; and iii) separate handling of pose heatmaps and visual features. Our approach aims to capture long-range motion dynamics in sparsely-labeled videos by modeling and aggregating temporal representations and motion context simultaneously. This enhances pose detection accuracy and reduces the need for dense annotations.

\section{Proposed Method}

\textbf{Problem formulation.} Following the top-down pose estimation paradigm, we initially utilize an object detector to extract the bounding box for each individual person in a video frame $I_{t}$. Each bounding box is then enlarged by 25\% to crop the same individual person in key frame $I_{t}$ and two auxiliary frames $I_{l}$ and $I_{r}$. $I_{l}$ and $I_{r}$ represent the key frame's left and right frames in the video sequence, respectively. Note that the time interval $T$ is set to 7, meaning there are 6 frames between the two auxiliary frames. In this manner, we obtain a cropped image sequence $\boldsymbol{\mathcal{S}}_{\boldsymbol{t}}^{\boldsymbol{i}} = \left\langle I_{l}^{i},  I_{t}^{i} , I_{r}^{i}\right\rangle$ for person \( i \). Given $\boldsymbol{\mathcal{S}}_{\boldsymbol{t}}^{\boldsymbol{i}}$, our goal is to estimate the pose in $I^{i}_{t}$.

\textbf{Method overview.} As shown in Figure~\ref{fig:framework}, our framework consists of three main components: a Representation Extraction module, a SpatioTemporal Representation Encoder (STRE) (Sec~\ref{sec:stre}), and a Pose Decoder (Sec~\ref{sec:decoder}). 
\textbf{(1)} Specifically, we first perform spatial feature extraction on the image sequence \(\boldsymbol{\mathcal{S}}_{\boldsymbol{t}}^{\boldsymbol{i}}\) to obtain a feature sequence $\boldsymbol{\mathcal{F}}_{\boldsymbol{t}}^{\boldsymbol{i}} = \left\langle F_{l}^{i}, F_{t}^{i} , F_{r}^{i}\right\rangle$, and then utilize a head to convert \(\boldsymbol{\mathcal{F}}_{\boldsymbol{t}}^{\boldsymbol{i}}\) into a keypoint heatmap sequence \(\boldsymbol{\mathcal{H}}_{\boldsymbol{t}}^{\boldsymbol{i}} = \left\langle H_{l}^{i}, H_{t}^{i} , H_{r}^{i}\right\rangle\). 
\textbf{(2)} We further feed $\boldsymbol{\mathcal{F}}_{\boldsymbol{t}}^{\boldsymbol{i}}$ and \(\boldsymbol{\mathcal{H}}_{\boldsymbol{t}}^{\boldsymbol{i}}\) into the TFF and TKS of the STRE, respectively, to implement feature fusion and keypoint merging, resulting in a fused feature \( \tilde{F_{t}^{i}} \) and a merged keypoint heatmaps \( \tilde{H_{t}^{i}} \). 
\textbf{(3)} Following this, we apply a novel modified sigmoid function to highlight the motion regions and perform a channel compression operation to obtain a dynamic-aware mask \( M_{t}^{i} \). Finally, STDA utilizes a cross-attention algorithm to aggregate \( \tilde{F_{t}^{i}} \), \( \tilde{H_{t}^{i}} \), and \( M_{t}^{i} \) into the final representation, which is then passed to a detection head to output the pose estimation \( \bar{H_{t}^{i}} \). In the following sections, we will elaborate on these components in detail.

\subsection{SpatioTemporal Representation Encoder}\label{sec:stre}

Leveraging the established excellence of Vision Transformers~\cite{dosovitskiy2020image, liu2021swin} (ViT) in extracting spatial features, we choose ViT as the backbone of our model to extract pose features from the input sequence $\boldsymbol{\mathcal{S}}_{\boldsymbol{t}}^{\boldsymbol{i}}$. However, we encounter two significant challenges: 

(1) While ViT is adept at capturing global spatial dependencies, it struggles with visual tasks such as video pose estimation, which require precise localization of keypoints and often involve subtle local pose variations within videos. To overcome this limitation, we introduce a SpatioTemporal Representation Encoder (STRE) that models both temporal features and temporal pose heatmaps concurrently. The pose heatmaps provide essential local spatial information to accurately pinpoint the locations of individual joints, complemented by global semantic information from the features to effectively handle complex scenes, such as pose occlusion and blur. 

(2) Additionally, directly fusing these temporal representations can introduce a substantial amount of task-irrelevant information, particularly because the temporal distance between auxiliary frames and key frames in sparsely-labeled videos can be significant. Drawing inspiration from prior works~\cite{liu2022temporal, hjelm2018learning, tian2021farewell}, we employ a mutual information objective to refine the process of temporal information extraction, ensuring that only relevant data contributes to pose estimation.

These innovations mark our contribution in enhancing the accuracy of video pose estimation, addressing both the challenge of precise localization and the effective integration of temporal dynamics.

\textbf{Temporal Feature Fusion (TFF) module.}\label{TFF}
We feed the $\boldsymbol{\mathcal{F}}_{\boldsymbol{t}}^{\boldsymbol{i}} = \left\langle F_{l}^{i}, F_{t}^{i} , F_{r}^{i}\right\rangle$ obtained from the backbone into the Temporal Feature Fusion module to output \(\tilde{F_{t}^{i}}\). Specifically, we first add a new learnable position embedding $PE$ to each visual feature and concatenate all the features together. We then feed them into cascaded Transformer blocks. Each block contains a multi-head self-attention layer and a feed-forward neural network. Finally, a Multilayer Perceptron is applied to aggregate all the encoded features to obtain \(\tilde{F_{t}^{i}}\). In summary, the Temporal Feature Fusion module is capable of effectively aggregating multi-frame temporal features, which markedly enhances the extraction of spatiotemporal information, making STDPose superior to existing methods that perform feature alignment~\cite{liu2022temporal} and representation difference learning~\cite{feng2023mutual}.

\textbf{Temporal Keypoints Synthesis (TKS) module.}\label{TKS}
While there is an inherent spatial correlation among the adjacent joints of the human body, the temporal trajectories of each individual joint posses a degree of independence~\cite{he2024video}. Thus, given a keypoint heatmap sequence \(\boldsymbol{\mathcal{H}}_{\boldsymbol{t}}^{\boldsymbol{i}}=\left\langle H_{l}^{i}, H_{t}^{i} , H_{r}^{i}\right\rangle\), we first merge each keypoint temporally using convolutional blocks, then merge all keypoints in the spatial dimension, and finally concatenate them with the sequence $\boldsymbol{\mathcal{H}}_{\boldsymbol{t}}^{\boldsymbol{i}}$ for the final synthesis to obtain the merged heatmaps \( \tilde{H_{t}^{i}} \). By such, our framework is able to learn spatiotemporal poses by synthesizing temporal keypoint heatmaps stepwise from different dimensions, which is a significant advancement over the previous approaches that involved the simplistic aggregation of pose information~\cite{liu2021deep, bertasius2019learning}.

\textbf{Mutual Information (MI) objective.}\label{MI} 
Directly modeling temporal pose and feature yields a significant amount of task-irrelevant clues in sparsely-labeled videos. To effectively extract temporal information, we introduce a mutual information (MI) objective. Within this framework, our main objective for learning effective temporal features and temporal poses can be formulated as:
\begin{equation} \label{eq_mi2}
\begin{aligned}
\max \left[\mathcal{I}\left(y_{t}^{i} ;  \tilde{F_{t}^{i}} \mid F_{t}^{i}\right)+\mathcal{I}\left(y_{t}^{i} ; \tilde{H_{t}^{i}} \mid H_{t}^{i}\right)\right],
\end{aligned}
\end{equation}
where \(y_{t}^{i}\) denotes the pose label. The terms $\mathcal{I}\left(y_{t}^{i} ;  \tilde{F_{t}^{i}} \mid F_{t}^{i}\right)$ and $\mathcal{I}\left(y_{t}^{i} ; \tilde{H_{t}^{i}} \mid H_{t}^{i}\right)$ each represent the measure of task-relevant information contained within the fused feature \(\tilde{F_{t}^{i}}\) and the merged heatmaps \(\tilde{H_{t}^{i}}\), respectively, that is in addition to the information already present in $F_{t}^{i}$ and $H_{t}^{i}$. Due to the difficulty in calculating mutual information~\cite{liu2022temporal}, we have simplified the formula, as detailed in Appendix. The proposed MI objective $\mathcal{L}_{\mathrm{MI}}$ is as follows:
\begin{equation}\label{eq_mi5}
\begin{aligned}
\mathcal{L}_{\mathrm{MI}}  = -\alpha \cdot \mathcal{I}\left(y_{t}^{i} ;  \tilde{F_{t}^{i}} \mid F_{t}^{i}\right) -  \beta \cdot \mathcal{I}\left(y_{t}^{i} ; \tilde{H_{t}^{i}} \mid H_{t}^{i}\right),
\end{aligned}
\end{equation}
where $\alpha$ and $\beta$ are hyperparameters to balance the ratio of different MI loss terms.

\begin{table*}[t] 
    \centering
    
    \resizebox{0.92\textwidth}{!}{
    \begin{tabular}{c|c|ccccccc|c}
        \hline
        Dataset & Method & Head & Shoulder & Elbow & Wrist & Hip & Knee & Ankle & Mean \\
        \hline
         \multirow{6}{*}{PoseTrack17 Val Set}  & Farneback~\cite{farneback2003two} & 76.5 & 82.3 & 74.3 & 69.2 & 80.8 & 74.8 & 70.1 & 75.5 \\
                            & SimpleBaseline~\cite{xiao2018simple} & 87.3 & 88.4 & 83.6 & 77.6 & 83.2 & 78.4 & 73.7 &  82.0\\
                            & FlowNet2~\cite{ilg2017flownet} & 82.7 & 91.0 & 83.8 & 78.4 & 89.7 & 83.6 & 78.1 & 83.8 \\
                            & DCPose~\cite{liu2021deep} & 91.2 & 90.8 & 88.4 & 84.3 & 87.7 & 86.7 & 83.2 & 87.7 \\                            
                            & PoseWarper~\cite{bertasius2019learning} & 86.0 & 92.7 & 89.5 & 86.0 & 91.5 & 89.1 & 86.6 & 88.7 \\
                            &  \textbf{STDPose (Ours)} & \textbf{92.7} & \textbf{93.1} & \textbf{91.4} & \textbf{88.1} & \textbf{91.9} & \textbf{90.3} & \textbf{88.1} & \textbf{90.9} \\
                         
        \hline

        \multirow{4}{*}{PoseTrack18 Val Set} & SimpleBaseline~\cite{xiao2018simple} & 82.7 & 80.1 & 72.7 & 66.1 & 72.1 & 69.7 & 65.6 & 73.4 \\
                            & PoseWarper~\cite{bertasius2019learning} & 87.0 & 88.5 & 84.8 & 80.4 & 81.4 & 82.1 & 79.9 & 83.7 \\
                            & DCPose~\cite{liu2021deep} & 88.9 & 89.0 & 85.6 & 81.8 & 84.8 & 82.7 & 80.3 & 85.0 \\
                            & \textbf{STDPose (Ours)} & \textbf{90.4} & \textbf{91.6} & \textbf{87.6} & \textbf{85.8} & \textbf{86.9} & \textbf{87.0} & \textbf{85.8} & \textbf{88.0} \\

        \hline
        \multirow{4}{*}{PoseTrack21 Val Set} & SimpleBaseline~\cite{xiao2018simple} & 82.2 & 77.6 & 71.1 & 64.1 & 68.6 & 65.0 & 59.6 & 70.6 \\
                            & PoseWarper~\cite{bertasius2019learning} & 88.6 & 87.0 & 83.3 & 79.2 & 80.7 & 80.4 & 77.3 & 82.8 \\
                            & DCPose~\cite{liu2021deep} & 88.1 & 87.3 & 82.8 & 79.7 & 83.0 & 79.6 & 78.2 & 83.1 \\
                            & \textbf{STDPose (Ours)} & \textbf{91.5} & \textbf{90.4} & \textbf{87.2} & \textbf{83.3} & \textbf{85.9} & \textbf{85.2} & \textbf{83.2} & \textbf{86.9} \\

        \hline
    \end{tabular}
    }
    \caption{The results of video pose propagation on PoseTrack2017~\cite{iqbal2017posetrack}, PoseTrack2018~\cite{andriluka2018posetrack}, and PoseTrack2021~\cite{doering2022posetrack21} datasets. Same as PoseWarper~\cite{bertasius2019learning}, all time intervals $T$ are set to 7, \textit{i.e.}, pose annotations are given every 7 frames. The evaluation metric is mean Average Precision (mAP).} 
    \label{table:compare_sparse}
\end{table*}

\subsection{Pose Decoder}\label{sec:decoder}

Instead of merely employing convolutional operations to combine keypoint heatmaps with features, which often leads to suboptimal outcomes due to a narrow focus on local interactions, we propose a more innovative strategy. Drawing on insights from recent multimodal research \cite{rombach2022high}, we formulate the integration of keypoint heatmaps and visual features as a quasi-multimodal task using a cross-attention algorithm. This algorithm excels at both local positioning and global searching through its robust similarity computation mechanism, allowing the framework to capture broader spatial interdependencies that are crucial for accurate pose estimation in complex video scenarios.
 
Moreover, temporal inconsistencies across frames are frequently encountered in sparsely-labeled videos. To tackle this challenge, we introduce a novel concept: the Dynamic-Aware Mask (DAM). This mask is designed to learn motion context effectively, thereby highlighting areas of movement and enhancing the reliability and precision of our framework. This innovative approach allows the framework to capture long-range motion clues, thus enabling a more nuanced understanding of pose dynamics, especially in challenging scenarios prone to occlusion or blur.

\textbf{Dynamic-Aware Mask (DAM) generation.}\label{DAM}
We are the first to propose the Dynamic-Aware Mask (DAM), designed to capture subtle motion clues in complex spatiotemporal interaction scenarios and offer a nuanced comprehension of pose dynamics. We first obtain forward and backward pose residuals through heatmap subtraction, then activate motion areas using a modified sigmoid function on each channel. We then compress the channels into a single channel and add the forward and backward feature maps with weighting. Finally, we utilize a softmax function to derive the mask \( M_{t}^{i} \). The proposed modified sigmoid function as follows:
\begin{equation}\label{eq_sigmoid}
\begin{aligned}
\boldsymbol{\mathcal{S}ig}_{m}(x, k, \theta) = \frac{1}{1 + e^{-k \cdot (\lvert x \rvert - \theta)}} ,
\end{aligned}
\end{equation} 
where $\boldsymbol{\mathcal{S}ig}_{m}(\cdot)$ is the modified sigmoid function. \(x\) is the input of the function. $k$ is a positive slope parameter that controls the steepness of the function. $\theta$ is a threshold parameter that determines the value of the function when $x=0$. We aim to keep the value at $x=0$ sufficiently small but not zero. $\lvert x \rvert$ represents the absolute value of $x$, ensuring that the function responds identically to both positive and negative values of $x$. This is because negative pose residual values also represent significant local spatial changes, which a regular sigmoid function would suppress. $e$ represents the mathematical constant. We empirically set $k$ and $\theta$ to 1.5 and 0.5, respectively. The mask generated by utilizing this sigmoid function can learn local spatial pose differences, thereby capturing subtle motion dynamics and enhancing the robustness of our framework.

\textbf{SpatioTemporal Dynamics Aggregation (STDA) module.}\label{STDA}
Given $\tilde{F_{t}^{i}}$, $\tilde{H_{t}^{i}}$, \( M_{t}^{i} \), the goal of STDA is to aggregate them and output the final heatmaps \( \bar{H_{t}^{i}} \). The process starts by performing a dot product between $\tilde{H_{t}^{i}}$ and \( M_{t}^{i} \) to obtain the masked heatmaps. These masked heatmaps are then concatenated with $\tilde{H_{t}^{i}}$ and fed into a patch embedding layer that embeds the heatmaps into tokens, also adding position embedding to get $\hat{H_{t}^{i}}$. Both $\tilde{F_{t}^{i}}$ and $\hat{H_{t}^{i}}$ are then fed into Pose Aggregation (PA) blocks, each consisting of a self-attention layer, a cross-attention layer, and a Feed-forward neural network. We also insert a LayerNorm operation before each layer. Ultimately, a classic pose detection head is engaged to upsample the features emanating from the last block, yielding the final predicted heatmaps \( \bar{H_{t}^{i}} \). Overall, the SpatioTemporal Dynamics Aggregation module integrates spatiotemporal representations and performs global-local learning of spatiotemporal dynamics to enhance the performance of pose detection using a cross-attention algorithm instead of deformable convolutions~\cite{zhu2019deformable, dai2017deformable}.

\subsection{Loss Functions}\label{sec:loss}

Our loss functions consist of two portions. (1) We employ the common pose heatmap loss $\mathcal{L}_{\mathrm{H}}$ to supervise the learning of the final pose heatmaps $\bar{H_{t}^{i}}$:
\begin{equation}\label{eq_pose_loss}
\begin{aligned}
 \mathcal{L}_{\mathrm{H}} = \left\| \bar{H_{t}^{i}} - G_{t}^{i}\right\|_{2}^{2},
\end{aligned}
\end{equation}
where $G_{t}^{i}$ denotes the ground-truth heatmaps. (2) Furthermore, We adopt the MI loss $\mathcal{L}_{\mathrm{MI}}$ from Eq.~\ref{eq_mi5} to supervise the extraction of temporal information. The total loss function $\mathcal{L}_{\mathrm{total}}$ is given by:
\begin{equation}\label{eq_total_loss}
\begin{aligned}
\mathcal{L}_{\mathrm{total}} = \mathcal{L}_{\mathrm{H}} + \mathcal{L}_{\mathrm{MI}}.
\end{aligned}
\end{equation}

\begin{table*}[t] 
    \centering
    
    \resizebox{0.95\textwidth}{!}{
    \begin{tabular}{c| c|ccccccc|c}
        \hline
        Dataset & Method & Head & Shoulder & Elbow & Wrist & Hip & Knee & Ankle & Mean \\
        \hline
         \multirow{9}{*}{PoseTrack17 Val Set} & PoseFlow~\cite{xiu2018pose} & 66.7 & 73.3 & 68.3 & 61.1 & 67.5 & 67.0 & 61.3 & 66.5 \\
                            & PoseWarper~\cite{bertasius2019learning}  &  81.4 &88.3 &83.9& 78.0& 82.4 &80.5 &73.6 & 81.2  \\
                            & DCPose~\cite{liu2021deep} & 88.0 & 88.7 & 84.1 & 78.4 &  83.0 & 81.4 & 74.2 & 82.8  \\
                            & FAMI-Pose~\cite{liu2022temporal} & 89.6  & 90.1 & 86.3 & 80.0 & 84.6 & 83.4 & 77.0 & 84.8   \\
                            & DSTA~\cite{he2024video} & 89.3  & 90.6 & 87.3 & 82.6 & 84.5 & 85.1 & 77.8 & 85.6   \\
                            & TDMI~\cite{feng2023mutual}  & \textbf{90.6} & 91.0 & 87.2 & 81.5 &  85.2 & 84.5 & 78.7 & 85.9 \\
                            & DiffPose~\cite{feng2023diffpose} & 89.0 & 91.2 & 87.4 & 83.5 & 85.5 & 87.2 & 80.2 & 86.4  \\
                            & \textbf{STDPose (Ours)} & 89.7 & \textbf{91.3} & \textbf{88.5} & \textbf{84.7} & \textbf{88.7} & \textbf{87.9} &  \textbf{80.6} & \textbf{87.4} \\

        \hline

        \multirow{8}{*}{PoseTrack18 Val Set}  & PoseWarper~\cite{bertasius2019learning}  & 79.9 & 86.3  &82.4  & 77.5 & 79.8  &78.8 & 73.2  &79.7 \\
                            & DCPose~\cite{liu2021deep}  &84.0& 86.6& 82.7& 78.0& 80.4 &79.3 &  73.8& 80.9 \\
                            & FAMI-Pose~\cite{liu2022temporal}  &85.5& 87.7 &84.2& 79.2& 81.4 &81.1 &  74.9& 82.2  \\
                            & DiffPose~\cite{feng2023diffpose}  &85.0 &87.7& 84.3 &81.5 &81.4& 82.9 &  77.6 &83.0 \\
                            & DSTA~\cite{he2024video} & 85.9  & 88.8 & 85.0 & 81.1 & 81.5 & 83.0 &  77.4 & 83.4   \\
                            & TDMI~\cite{feng2023mutual}  &\textbf{86.7}& \textbf{88.9} &85.4 &80.6 &82.4& 82.1&  77.6& 83.6  \\       
                            & \textbf{STDPose (Ours)} & 84.9 & 88.3 & \textbf{85.5} & \textbf{82.3} & \textbf{85.9} & \textbf{84.9} &  \textbf{79.6} &  \textbf{84.5} \\

        \hline
        \multirow{6}{*}{PoseTrack21 Val Set} & DCPose~\cite{liu2021deep} &83.2 &84.7& 82.3 &78.1 &80.3 &79.2 & 73.5& 80.5  \\
                            & FAMI-Pose~\cite{liu2022temporal}  &83.3 &85.4 &82.9 &78.6 &81.3 &80.5 & 75.3 &81.2 \\
                            & DiffPose~\cite{feng2023diffpose}   &84.7& 85.6 &83.6 &80.8 &81.4& 83.5 &   \textbf{80.0} &82.9 \\
                            & DSTA~\cite{he2024video} & \textbf{87.5}  & 87.0 & 84.2 & 81.4 & 82.3 & 82.5 &  77.7 & 83.5   \\
                            & TDMI~\cite{feng2023mutual}  &86.8 &87.4 &85.1 &81.4 &83.8 &82.7 &  78.0 &83.8 \\
                            & \textbf{STDPose (Ours)} & 84.4 &  \textbf{88.5} & \textbf{85.3} & \textbf{82.3} & \textbf{85.5} & \textbf{84.1} &  \textbf{80.0} &  \textbf{84.3} \\

        \hline
    \end{tabular}
    }
    \caption{Comparisons with the state-of-the-art methods for video pose estimation on the validation sets of the PoseTrack2017~\cite{iqbal2017posetrack}, PoseTrack2018~\cite{andriluka2018posetrack}, and PoseTrack2021~\cite{doering2022posetrack21} datasets. Note that during training, we aggregate temporal information from neighboring frames (\textit{i.e.}, one frame to the left and one to the right), and during inference, the pose labels of neighboring frames are not provided.} 
    \label{table:compare_no_sparse}
\end{table*}

\section{Experiments}

\subsection{Experimental Settings}
We carried out thorough evaluations for video pose propagation and video pose estimation tasks on three popular benchmarks: PoseTrack2017~\cite{iqbal2017posetrack}, PoseTrack2018~\cite{andriluka2018posetrack}, and PoseTrack21~\cite{doering2022posetrack21}. The videos in these datasets feature diverse challenges, such as crowded scenes and rapid movements. The input image size is 256$\times$192. We utilize a standard Vision Transformer~\cite{dosovitskiy2020image} pretrained on the COCO dataset~\cite{lin2014microsoft} as the backbone network of our STDPose framework. We set the parameters $\alpha$ to 0.1 and $\beta$ to 0.01 in Eq.~\ref{eq_mi5}, and have not densely tuned them. We evaluate our model using the standard pose estimation metric, average precision (\textbf{AP}), by initially calculating the AP for each joint and subsequently deriving the model's overall performance through the mean average precision (\textbf{mAP}) across all joints.

\subsection{Comparison on Pose Propagation}

We apply our model to the video pose propagation task, \textit{i.e.}, propagating poses across time from a few labeled frames. Specifically, during training, every 7th frame of the training videos acts as an auxiliary frame (\textit{i.e.}, there are 6 key frames between each pair of auxiliary frames) with no pose annotation. Subsequently, during inference, we provide pose annotations for the auxiliary frames to facilitate pose propagation from labeled (auxiliary) frames to all unlabeled frames. We compare our model with several state-of-the-art methods, including DCPose~\cite{liu2021deep}, and PoseWarper~\cite{bertasius2019learning}, among others. We test the performance of applying DCPose to sparsely-labeled videos based on its open-source release. Regrettably, a direct quantitative comparison with recent studies \cite{liu2022temporal, feng2023mutual} is not possible due to their inherent limitations in aggregating pose labels and propagating poses within sparsely-labeled videos. The experimental results on three benchmark datasets are in Table~\ref{table:compare_sparse}.

\textbf{PoseTrack2017.} 
The proposed STDPose consistently surpasses existing methods, achieving an mAP of 90.9. Our model achieves a 2.2 mAP gain over the previous state-of-the-art approach PoseWarper~\cite{bertasius2019learning}. Especially, we obtain promising improvements for the more challenging joints (\textit{i.e.}, wrist, ankle), with an mAP of 88.1 ($\uparrow$ 2.1) for wrists and an mAP of 88.1 ($\uparrow$ 1.5) for ankles.

\textbf{PoseTrack2018.}
STDPose surpasses the previous state-of-the-art method DCPose~\cite{liu2021deep}, attaining an mAP of 88.0 ($\uparrow$ 3.0), with an mAP of 85.8, 87.0, and 85.8 for the wrist, knee, and ankle, respectively. Our model outperforms DCPose and PoseWarper~\cite{bertasius2019learning}, which only process heatmaps, demonstrating the contribution of concurrently integrating both heatmaps and features in our approach.

\textbf{PoseTrack2021.}  
STDPose achieves a 86.9 mAP---3.8 higher than DCPose~\cite{liu2021deep}. It greatly surpasses other methods on wrists and ankles, showcasing its effectiveness in tackling challenging scenarios where these joints are often blurred or occluded due to pose occlusion and rapid movement.

\subsection{Comparison on Pose Estimation}
 
\textbf{Pose estimation trained on full video annotation.} Unlike pose propagation, during pose estimation training, the auxiliary frames are provided from neighboring frames, and no auxiliary frame pose labels are required during inference. Notably, similar to DCPose \cite{liu2021deep}, we only require two auxiliary frames in training, whereas recent methods like \cite{he2024video,feng2023mutual} require four.

As shown in Table~\ref{table:compare_no_sparse}, STDPose outperforms all comparing methods across three datasets, achieving a 87.4 mAP on PoseTrack2017 (1.0 higher than DiffPose~\cite{feng2023diffpose}) and a 0.9 mAP improvement on PoseTrack2018 over TDMI~\cite{feng2023mutual} (with a significant 2.0 mAP increase for the ankle joint). STDPose also tops on the PoseTrack2021 dataset with a 84.3 mAP. Our method excels in detecting challenging joints like wrists and ankles. This is attributed to the combination of the global receptive field provided by the attention mechanism and the dynamic awareness facilitated by the DAM. The superior performance in complex interaction scenarios demonstrates the robustness of our model.

\textbf{Pose estimation trained with pseudo-labels generated by pose propagation.} We further demonstrate the effectiveness of our pose propagation model in enhancing pose estimation on sparsely-labeled videos. Specifically, we train our model using a combination of manual annotations and pseudo-labels generated by the pose propagation model on the PoseTrack2017 training set. By varying parameter $T$, we control the proportion of manually-labeled frames, with $T$=2 indicating a 50/50 split. We then evaluate the pose estimation performance on PoseTrack2017 validation set.

As shown in Table \ref{table:pseudo}, pseudo-labels generated from pose propagation significantly improves pose estimation when dealing with sparsely-labeled videos. Our model achieves 84.3 mAP at $T$=4, close to FAMI-Pose~\cite{liu2022temporal}. Notably, at $T$=2, \textbf{our model excels over FAMI-Pose \cite{liu2022temporal}, achieving 85.2 mAP with only 50\% of the manually-labeled frames, demonstrating superior performance with only half the labeled data}. Our model also outperforms FAMI-Pose in ankle joint detection accuracy at $T$=4. These results demonstrate that high accuracy can be achieved with minimal labeled data, reducing the need for large, densely annotated datasets. For brevity, we compare only with the classic method FAMI-Pose \cite{liu2022temporal} in Table \ref{table:pseudo}, while comparisons with other state-of-the-art methods are available in Table \ref{table:compare_no_sparse}.

\begin{table*}[h] 
    \centering
    
    \resizebox{0.96\linewidth}{!}{
    \begin{tabular}{c|c|c|ccccccc|c}
        \hline
         Model & $T$ & Labeled Frame Ratio  & Head & Shoulder & Elbow & Wrist & Hip & Knee & Ankle & Mean \\
        \hline
         FAMI-Pose~\cite{liu2022temporal} &  - & \textbf{100\%}  & 89.6  & 90.1 & 86.3 & 80.0 & 84.6 & 83.4 & 77.0 & \textbf{84.8}   \\
        \hline
         \multirow{4}{*}{STDPose (Ours)} & $T=7$ & 16.7\% & 87.1 & 85.9 & 82.8 & 79.6 & 83.1 & 82.0 & 76.3 & 82.4 ($\downarrow$ 2.4) \\
          & $T=4$ & \textbf{26.7\%} &   88.3 & 87.8 & 84.1 & 81.5 & 85.3 & 84.2 & 79.0 & \textbf{84.3}  ($\downarrow$ 0.5)\\
          & $T=2$ &  \textbf{50.0}\%  & 88.9 & 89.0 & 85.4 & 82.6 & 86.6 & 84.9 & 79.6 & \textbf{85.2}  ($\uparrow$ 0.4)\\  
          &   - &  100.0\%  &  89.7 & 91.3 & 88.5 & 84.7 & 88.7 & 87.9 &  80.6 & 87.4 ($\uparrow$ 2.6)\\ 
        \hline
    \end{tabular}
    }
    \caption{The results of the pose estimation model trained with pseudo-labels generated by a pose propagation model at different $T$ on the PoseTrack2017 validation set.}\label{table:pseudo}
\end{table*}

\begin{table*}[t] \small
\resizebox{!}{2pt}{\checkmark}
\centering
\resizebox{0.80\textwidth}{!}{
\begin{tabular}{ccccccccc}
\midrule
\multirow{2}{*}{Methods} & \multirow{2}{*}{ $T$} & Labeled Frame  & \multicolumn{5}{c}{ Components} &   \multirow{2}{*}{ mAP (\%)}  \\ 
\cmidrule{4-8}
              & & Ratio &  TFF &  TKS                   &  DAM &  STDA &   MI      \\
\midrule
(a) & - & 0.0\% & & & & & & 86.8  \\
(b) & 7 & 16.7\% & \checkmark & & & & & 88.4 ($\uparrow$ 1.6) \\
(c) & 7 & 16.7\% &   & \checkmark& &  &  & 88.6 ($\uparrow$ 1.8)\\
(d) & 7  & 16.7\% &  \checkmark & \checkmark &  & \checkmark &  & 89.6 ($\uparrow$ 2.8)\\
(e) &  7  & 16.7\% &  \checkmark &\checkmark & \checkmark & \checkmark &  & 90.3 ($\uparrow$ 3.5)\\
\midrule
(f) & 7 & 16.7\% &  \checkmark  & \checkmark &\checkmark & \checkmark & \checkmark & 90.9 ($\uparrow$ 4.1)\\

\midrule
\multirow{6}{*}{STDPose (Ours)} &  15 &  10.0\% &   \checkmark & \checkmark & \checkmark & \checkmark & \checkmark  &   83.6 \\
 &  9 &   13.3\% &  \checkmark & \checkmark & \checkmark & \checkmark & \checkmark  &  90.1   \\
 &  5 &    23.3\% &   \checkmark & \checkmark & \checkmark & \checkmark & \checkmark  &  92.5   \\
 &  3 &    36.7\% &   \checkmark & \checkmark & \checkmark & \checkmark & \checkmark  &  96.8   \\
 &  2 &    50.0\% &   \checkmark & \checkmark & \checkmark & \checkmark & \checkmark  &  \textbf{98.1}   \\
\midrule
\end{tabular}
}
\caption{The upper half of the table presents an ablation study on various components within our STDPose method, conducted on the PoseTrack2017~\cite{iqbal2017posetrack} validation set for the pose propagation task. The lower half illustrates the impact of different time intervals $T$ on performance.} 
\label{table:ablation}
\end{table*}

\subsection{Ablation Study}
We first perform ablation experiments to examine the influence of each component in our proposed method STDPose for pose propagation task. We also examine the performance changes in pose propagation as the time interval $T$ increases or decreases. All the ablation studies are conducted on the PoseTrack2017 validation set.

\textbf{Study on components of STDPose.} 
We conduct a comprehensive evaluation of each component in our proposed STDPose framework, presenting the quantitative results in Table~\ref{table:ablation}. \textbf{(a)} As a baseline, we incorporate a pose detection head after the ViT backbone. \textbf{(b)} We add only the Temporal Feature Fusion (TFF) module. \textbf{(c)} We include only the Temporal Keypoints Synthesis (TKS) module. \textbf{(d)} Both the TFF and TKS modules are added, utilizing the SpatioTemporal Dynamics Aggregation (STDA) module to aggregate spatiotemporal representations. \textbf{(e)} The Dynamic-Aware Mask (DAM) is introduced to effectively learn motion context. \textbf{(f)} We incorporate the Mutual Information objective into the loss function to form the complete STDPose. Results reveal that combining TFF and TKS offers superior performance compared to using each module individually. Additionally, incorporating DAM results in a performance improvement of 0.7 mAP. Lastly, the integration of MI allows our framework to achieve the highest performance, improving by 4.1 mAP over the baseline.

\textbf{Study on different time intervals.} 
Additionally, we assess the performance of our method across varying time intervals $T$, as detailed in Table~\ref{table:ablation}. Considering the average video contains about 30 frames, setting $T$ to 7 results in roughly 5 evenly distributed labeled frames per video. Increasing $T$ reduces the number of required labeled frames: at 
$T$=15 a video contains only 3 labeled frames, representing just 10\% of the total. However, this configuration leads to a significant drop in performance. Conversely, when $T$ is reduced to 2, equating the number of labeled and unlabeled frames, our method reaches an mAP of 98.1, nearly matching a 100\% ideal precision of manual annotation. This promising result highlights our method's ability to efficiently automate video annotation, achieving near-perfect accuracy with just 50\% of frames labeled manually.

\subsection{Discussion}

In summary, STDPose consistently excels in both pose propagation and pose estimation across all datasets, particularly in challenging scenarios with blurred or occluded joints like wrists and ankles. Our pose propagation model significantly boosts pose estimation on sparsely-labeled videos, achieving competitive performance with minimal labeled data. These results showcase STDPose's effectiveness in learning video spatiotemporal clues, not only enhancing performance but also tremendously reducing manual labeling efforts.

\section{Conclusion}
We introduce STDPose, a novel architecture for video pose propagation and pose estimation. STDPose innovatively models temporal features and pose heatmaps simultaneously, setting a new standard in the field. We also introduce a theoretical advancement, the Dynamic-Aware Mask, which is specifically designed to learn and interpret motion dynamics effectively, especially in challenging occasions such as occlusion and blur. STDPose pushes the edge in performance on three benchmark datasets, while also reducing the need for extensive manual video annotations. 


\section*{Acknowledgments} 
This work is supported by the National Natural Science Foundation of China (No. 62372402), and the Key R\&D Program of Zhejiang Province (No. 2023C01217).

\bibliography{aaai25}

\begin{thebibliography}{45}
\providecommand{\natexlab}[1]{#1}

\bibitem[{Andriluka et~al.(2018)Andriluka, Iqbal, Insafutdinov, Pishchulin, Milan, Gall, and Schiele}]{andriluka2018posetrack}
Andriluka, M.; Iqbal, U.; Insafutdinov, E.; Pishchulin, L.; Milan, A.; Gall, J.; and Schiele, B. 2018.
\newblock Posetrack: A benchmark for human pose estimation and tracking.
\newblock In \emph{Proceedings of the IEEE conference on computer vision and pattern recognition}, 5167--5176.

\bibitem[{Artacho and Savakis(2020)}]{artacho2020unipose}
Artacho, B.; and Savakis, A. 2020.
\newblock Unipose: Unified human pose estimation in single images and videos.
\newblock In \emph{Proceedings of the IEEE/CVF conference on computer vision and pattern recognition}, 7035--7044.

\bibitem[{Bertasius et~al.(2019)Bertasius, Feichtenhofer, Tran, Shi, and Torresani}]{bertasius2019learning}
Bertasius, G.; Feichtenhofer, C.; Tran, D.; Shi, J.; and Torresani, L. 2019.
\newblock Learning temporal pose estimation from sparsely-labeled videos.
\newblock \emph{Advances in neural information processing systems}, 32.

\bibitem[{Chen et~al.(2023)Chen, Wang, Zhou, Qiao, and Dong}]{chen2023activating}
Chen, X.; Wang, X.; Zhou, J.; Qiao, Y.; and Dong, C. 2023.
\newblock Activating more pixels in image super-resolution transformer.
\newblock In \emph{Proceedings of the IEEE/CVF conference on computer vision and pattern recognition}, 22367--22377.

\bibitem[{Dai et~al.(2017)Dai, Qi, Xiong, Li, Zhang, Hu, and Wei}]{dai2017deformable}
Dai, J.; Qi, H.; Xiong, Y.; Li, Y.; Zhang, G.; Hu, H.; and Wei, Y. 2017.
\newblock Deformable convolutional networks.
\newblock In \emph{Proceedings of the IEEE international conference on computer vision}, 764--773.

\bibitem[{Deng et~al.(2009)Deng, Dong, Socher, Li, Li, and Fei-Fei}]{deng2009imagenet}
Deng, J.; Dong, W.; Socher, R.; Li, L.-J.; Li, K.; and Fei-Fei, L. 2009.
\newblock Imagenet: A large-scale hierarchical image database.
\newblock In \emph{2009 IEEE conference on computer vision and pattern recognition}, 248--255. Ieee.

\bibitem[{Doering et~al.(2022)Doering, Chen, Zhang, Schiele, and Gall}]{doering2022posetrack21}
Doering, A.; Chen, D.; Zhang, S.; Schiele, B.; and Gall, J. 2022.
\newblock Posetrack21: A dataset for person search, multi-object tracking and multi-person pose tracking.
\newblock In \emph{Proceedings of the IEEE/CVF Conference on Computer Vision and Pattern Recognition}, 20963--20972.

\bibitem[{Dosovitskiy et~al.(2020)Dosovitskiy, Beyer, Kolesnikov, Weissenborn, Zhai, Unterthiner, Dehghani, Minderer, Heigold, Gelly et~al.}]{dosovitskiy2020image}
Dosovitskiy, A.; Beyer, L.; Kolesnikov, A.; Weissenborn, D.; Zhai, X.; Unterthiner, T.; Dehghani, M.; Minderer, M.; Heigold, G.; Gelly, S.; et~al. 2020.
\newblock An image is worth 16x16 words: Transformers for image recognition at scale.
\newblock \emph{arXiv preprint arXiv:2010.11929}.

\bibitem[{Farneb{\"a}ck(2003)}]{farneback2003two}
Farneb{\"a}ck, G. 2003.
\newblock Two-frame motion estimation based on polynomial expansion.
\newblock In \emph{Image Analysis: 13th Scandinavian Conference, SCIA 2003 Halmstad, Sweden, June 29--July 2, 2003 Proceedings 13}, 363--370. Springer.

\bibitem[{Feng et~al.(2023{\natexlab{a}})Feng, Gao, Ma, Tse, and Chang}]{feng2023mutual}
Feng, R.; Gao, Y.; Ma, X.; Tse, T. H.~E.; and Chang, H.~J. 2023{\natexlab{a}}.
\newblock Mutual information-based temporal difference learning for human pose estimation in video.
\newblock In \emph{Proceedings of the IEEE/CVF Conference on Computer Vision and Pattern Recognition}, 17131--17141.

\bibitem[{Feng et~al.(2023{\natexlab{b}})Feng, Gao, Tse, Ma, and Chang}]{feng2023diffpose}
Feng, R.; Gao, Y.; Tse, T. H.~E.; Ma, X.; and Chang, H.~J. 2023{\natexlab{b}}.
\newblock DiffPose: SpatioTemporal diffusion model for video-based human pose estimation.
\newblock In \emph{Proceedings of the IEEE/CVF International Conference on Computer Vision}, 14861--14872.

\bibitem[{He and Yang(2024)}]{he2024video}
He, J.; and Yang, W. 2024.
\newblock Video-Based Human Pose Regression via Decoupled Space-Time Aggregation.
\newblock \emph{arXiv preprint arXiv:2403.19926}.

\bibitem[{Hjelm et~al.(2018)Hjelm, Fedorov, Lavoie-Marchildon, Grewal, Bachman, Trischler, and Bengio}]{hjelm2018learning}
Hjelm, R.~D.; Fedorov, A.; Lavoie-Marchildon, S.; Grewal, K.; Bachman, P.; Trischler, A.; and Bengio, Y. 2018.
\newblock Learning deep representations by mutual information estimation and maximization.
\newblock \emph{arXiv preprint arXiv:1808.06670}.

\bibitem[{Ilg et~al.(2017)Ilg, Mayer, Saikia, Keuper, Dosovitskiy, and Brox}]{ilg2017flownet}
Ilg, E.; Mayer, N.; Saikia, T.; Keuper, M.; Dosovitskiy, A.; and Brox, T. 2017.
\newblock Flownet 2.0: Evolution of optical flow estimation with deep networks.
\newblock In \emph{Proceedings of the IEEE conference on computer vision and pattern recognition}, 2462--2470.

\bibitem[{Iqbal, Milan, and Gall(2017)}]{iqbal2017posetrack}
Iqbal, U.; Milan, A.; and Gall, J. 2017.
\newblock Posetrack: Joint multi-person pose estimation and tracking.
\newblock In \emph{Proceedings of the IEEE Conference on Computer Vision and Pattern Recognition}, 2011--2020.

\bibitem[{Jin, Lee, and Lee(2022)}]{jin2022otpose}
Jin, K.-M.; Lee, G.-H.; and Lee, S.-W. 2022.
\newblock OTPose: occlusion-aware transformer for pose estimation in sparsely-labeled videos.
\newblock In \emph{2022 IEEE International Conference on Systems, Man, and Cybernetics (SMC)}, 3255--3260. IEEE.

\bibitem[{Kirillov et~al.(2023)Kirillov, Mintun, Ravi, Mao, Rolland, Gustafson, Xiao, Whitehead, Berg, Lo et~al.}]{kirillov2023segment}
Kirillov, A.; Mintun, E.; Ravi, N.; Mao, H.; Rolland, C.; Gustafson, L.; Xiao, T.; Whitehead, S.; Berg, A.~C.; Lo, W.-Y.; et~al. 2023.
\newblock Segment anything.
\newblock In \emph{Proceedings of the IEEE/CVF International Conference on Computer Vision}, 4015--4026.

\bibitem[{Lin et~al.(2014)Lin, Maire, Belongie, Hays, Perona, Ramanan, Doll{\'a}r, and Zitnick}]{lin2014microsoft}
Lin, T.-Y.; Maire, M.; Belongie, S.; Hays, J.; Perona, P.; Ramanan, D.; Doll{\'a}r, P.; and Zitnick, C.~L. 2014.
\newblock Microsoft coco: Common objects in context.
\newblock In \emph{Computer Vision--ECCV 2014: 13th European Conference, Zurich, Switzerland, September 6-12, 2014, Proceedings, Part V 13}, 740--755. Springer.

\bibitem[{Liu et~al.(2021{\natexlab{a}})Liu, Chen, Feng, Wu, Ji, Yang, and Wang}]{liu2021deep}
Liu, Z.; Chen, H.; Feng, R.; Wu, S.; Ji, S.; Yang, B.; and Wang, X. 2021{\natexlab{a}}.
\newblock Deep dual consecutive network for human pose estimation.
\newblock In \emph{Proceedings of the IEEE/CVF conference on computer vision and pattern recognition}, 525--534.

\bibitem[{Liu et~al.(2022{\natexlab{a}})Liu, Feng, Chen, Wu, Gao, Gao, and Wang}]{liu2022temporal}
Liu, Z.; Feng, R.; Chen, H.; Wu, S.; Gao, Y.; Gao, Y.; and Wang, X. 2022{\natexlab{a}}.
\newblock Temporal feature alignment and mutual information maximization for video-based human pose estimation.
\newblock In \emph{Proceedings of the IEEE/CVF conference on computer vision and pattern recognition}, 11006--11016.

\bibitem[{Liu et~al.(2021{\natexlab{b}})Liu, Lin, Cao, Hu, Wei, Zhang, Lin, and Guo}]{liu2021swin}
Liu, Z.; Lin, Y.; Cao, Y.; Hu, H.; Wei, Y.; Zhang, Z.; Lin, S.; and Guo, B. 2021{\natexlab{b}}.
\newblock Swin transformer: Hierarchical vision transformer using shifted windows.
\newblock In \emph{Proceedings of the IEEE/CVF international conference on computer vision}, 10012--10022.

\bibitem[{Liu et~al.(2022{\natexlab{b}})Liu, Mao, Wu, Feichtenhofer, Darrell, and Xie}]{liu2022convnet}
Liu, Z.; Mao, H.; Wu, C.-Y.; Feichtenhofer, C.; Darrell, T.; and Xie, S. 2022{\natexlab{b}}.
\newblock A convnet for the 2020s.
\newblock In \emph{Proceedings of the IEEE/CVF conference on computer vision and pattern recognition}, 11976--11986.

\bibitem[{Liu et~al.(2022{\natexlab{c}})Liu, Wu, Xu, Wang, Zhu, Wu, and Feng}]{liu2022copy}
Liu, Z.; Wu, S.; Xu, C.; Wang, X.; Zhu, L.; Wu, S.; and Feng, F. 2022{\natexlab{c}}.
\newblock Copy motion from one to another: Fake motion video generation.
\newblock \emph{arXiv preprint arXiv:2205.01373}.

\bibitem[{Newell, Yang, and Deng(2016)}]{newell2016stacked}
Newell, A.; Yang, K.; and Deng, J. 2016.
\newblock Stacked hourglass networks for human pose estimation.
\newblock In \emph{Computer Vision--ECCV 2016: 14th European Conference, Amsterdam, The Netherlands, October 11-14, 2016, Proceedings, Part VIII 14}, 483--499. Springer.

\bibitem[{Pfister, Charles, and Zisserman(2015)}]{pfister2015flowing}
Pfister, T.; Charles, J.; and Zisserman, A. 2015.
\newblock Flowing convnets for human pose estimation in videos.
\newblock In \emph{Proceedings of the IEEE international conference on computer vision}, 1913--1921.

\bibitem[{Rombach et~al.(2022)Rombach, Blattmann, Lorenz, Esser, and Ommer}]{rombach2022high}
Rombach, R.; Blattmann, A.; Lorenz, D.; Esser, P.; and Ommer, B. 2022.
\newblock High-resolution image synthesis with latent diffusion models.
\newblock In \emph{Proceedings of the IEEE/CVF conference on computer vision and pattern recognition}, 10684--10695.

\bibitem[{Sapp, Toshev, and Taskar(2010)}]{sapp2010cascaded}
Sapp, B.; Toshev, A.; and Taskar, B. 2010.
\newblock Cascaded models for articulated pose estimation.
\newblock In \emph{Computer Vision--ECCV 2010: 11th European Conference on Computer Vision, Heraklion, Crete, Greece, September 5-11, 2010, Proceedings, Part II 11}, 406--420. Springer.

\bibitem[{Schmidtke et~al.(2021)Schmidtke, Vlontzos, Ellershaw, Lukens, Arichi, and Kainz}]{schmidtke2021unsupervised}
Schmidtke, L.; Vlontzos, A.; Ellershaw, S.; Lukens, A.; Arichi, T.; and Kainz, B. 2021.
\newblock Unsupervised human pose estimation through transforming shape templates.
\newblock In \emph{Proceedings of the IEEE/CVF Conference on Computer Vision and Pattern Recognition}, 2484--2494.

\bibitem[{Shuai et~al.(2023)Shuai, Zhong, Wu, Lin, Wang, Ba, Liu, Cavallaro, and Ren}]{shuai2023locate}
Shuai, C.; Zhong, J.; Wu, S.; Lin, F.; Wang, Z.; Ba, Z.; Liu, Z.; Cavallaro, L.; and Ren, K. 2023.
\newblock Locate and verify: A two-stream network for improved deepfake detection.
\newblock In \emph{Proceedings of the 31st ACM International Conference on Multimedia}, 7131--7142.

\bibitem[{Su et~al.(2021)Su, Liu, Wu, Zhu, Yin, and Shen}]{su2021motion}
Su, P.; Liu, Z.; Wu, S.; Zhu, L.; Yin, Y.; and Shen, X. 2021.
\newblock Motion prediction via joint dependency modeling in phase space.
\newblock In \emph{Proceedings of the 29th ACM international conference on multimedia}, 713--721.

\bibitem[{Sun et~al.(2019)Sun, Xiao, Liu, and Wang}]{sun2019deep}
Sun, K.; Xiao, B.; Liu, D.; and Wang, J. 2019.
\newblock Deep high-resolution representation learning for human pose estimation.
\newblock In \emph{Proceedings of the IEEE/CVF conference on computer vision and pattern recognition}, 5693--5703.

\bibitem[{Tian et~al.(2021)Tian, Zhang, Lin, Qu, Xie, and Ma}]{tian2021farewell}
Tian, X.; Zhang, Z.; Lin, S.; Qu, Y.; Xie, Y.; and Ma, L. 2021.
\newblock Farewell to mutual information: Variational distillation for cross-modal person re-identification.
\newblock In \emph{Proceedings of the IEEE/CVF Conference on Computer Vision and Pattern Recognition}, 1522--1531.

\bibitem[{Tse et~al.(2022)Tse, Kim, Leonardis, and Chang}]{tse2022collaborative}
Tse, T. H.~E.; Kim, K.~I.; Leonardis, A.; and Chang, H.~J. 2022.
\newblock Collaborative learning for hand and object reconstruction with attention-guided graph convolution.
\newblock In \emph{Proceedings of the IEEE/CVF Conference on Computer Vision and Pattern Recognition}, 1664--1674.

\bibitem[{Vaswani et~al.(2017)Vaswani, Shazeer, Parmar, Uszkoreit, Jones, Gomez, Kaiser, and Polosukhin}]{vaswani2017attention}
Vaswani, A.; Shazeer, N.; Parmar, N.; Uszkoreit, J.; Jones, L.; Gomez, A.~N.; Kaiser, {\L}.; and Polosukhin, I. 2017.
\newblock Attention is all you need.
\newblock \emph{Advances in neural information processing systems}, 30.

\bibitem[{Wang, Tighe, and Modolo(2020)}]{wang2020combining}
Wang, M.; Tighe, J.; and Modolo, D. 2020.
\newblock Combining detection and tracking for human pose estimation in videos.
\newblock In \emph{Proceedings of the IEEE/CVF Conference on Computer Vision and Pattern Recognition}, 11088--11096.

\bibitem[{Wang et~al.(2022)Wang, Li, Li, He, Huang, Zhao, Zhang, Xu, Liu, Wang et~al.}]{wang2022internvideo}
Wang, Y.; Li, K.; Li, Y.; He, Y.; Huang, B.; Zhao, Z.; Zhang, H.; Xu, J.; Liu, Y.; Wang, Z.; et~al. 2022.
\newblock Internvideo: General video foundation models via generative and discriminative learning.
\newblock \emph{arXiv preprint arXiv:2212.03191}.

\bibitem[{Wei et~al.(2016)Wei, Ramakrishna, Kanade, and Sheikh}]{wei2016convolutional}
Wei, S.-E.; Ramakrishna, V.; Kanade, T.; and Sheikh, Y. 2016.
\newblock Convolutional pose machines.
\newblock In \emph{Proceedings of the IEEE conference on Computer Vision and Pattern Recognition}, 4724--4732.

\bibitem[{Wu et~al.(2024{\natexlab{a}})Wu, Chen, Yin, Hu, Feng, Jiao, Yang, and Liu}]{wu2024joint}
Wu, S.; Chen, H.; Yin, Y.; Hu, S.; Feng, R.; Jiao, Y.; Yang, Z.; and Liu, Z. 2024{\natexlab{a}}.
\newblock Joint-Motion Mutual Learning for Pose Estimation in Video.
\newblock In \emph{Proceedings of the 32nd ACM International Conference on Multimedia}, 8962--8971.

\bibitem[{Wu et~al.(2024{\natexlab{b}})Wu, Liu, Zhang, Zimmermann, Ba, Zhang, and Ren}]{wu2024pose}
Wu, S.; Liu, Z.; Zhang, B.; Zimmermann, R.; Ba, Z.; Zhang, X.; and Ren, K. 2024{\natexlab{b}}.
\newblock Do as I Do: Pose Guided Human Motion Copy.
\newblock \emph{IEEE Transactions on Dependable and Secure Computing}.

\bibitem[{Xiao, Wu, and Wei(2018)}]{xiao2018simple}
Xiao, B.; Wu, H.; and Wei, Y. 2018.
\newblock Simple baselines for human pose estimation and tracking.
\newblock In \emph{Proceedings of the European conference on computer vision (ECCV)}, 466--481.

\bibitem[{Xiu et~al.(2018)Xiu, Li, Wang, Fang, and Lu}]{xiu2018pose}
Xiu, Y.; Li, J.; Wang, H.; Fang, Y.; and Lu, C. 2018.
\newblock Pose Flow: Efficient online pose tracking.
\newblock \emph{arXiv preprint arXiv:1802.00977}.

\bibitem[{Xu et~al.(2022)Xu, Zhang, Zhang, and Tao}]{xu2022vitpose}
Xu, Y.; Zhang, J.; Zhang, Q.; and Tao, D. 2022.
\newblock Vitpose: Simple vision transformer baselines for human pose estimation.
\newblock \emph{Advances in Neural Information Processing Systems}, 35: 38571--38584.

\bibitem[{Yang et~al.(2023)Yang, Chen, Liu, Lyu, Zhang, Wu, Wang, and Ren}]{yang2023action}
Yang, Y.; Chen, H.; Liu, Z.; Lyu, Y.; Zhang, B.; Wu, S.; Wang, Z.; and Ren, K. 2023.
\newblock Action recognition with multi-stream motion modeling and mutual information maximization.
\newblock \emph{arXiv preprint arXiv:2306.07576}.

\bibitem[{Zhang et~al.(2009)Zhang, Li, Tong, Hu, Maybank, and Zhang}]{zhang2009efficient}
Zhang, X.; Li, C.; Tong, X.; Hu, W.; Maybank, S.; and Zhang, Y. 2009.
\newblock Efficient human pose estimation via parsing a tree structure based human model.
\newblock In \emph{2009 IEEE 12th International Conference on Computer Vision}, 1349--1356. IEEE.

\bibitem[{Zhu et~al.(2019)Zhu, Hu, Lin, and Dai}]{zhu2019deformable}
Zhu, X.; Hu, H.; Lin, S.; and Dai, J. 2019.
\newblock Deformable convnets v2: More deformable, better results.
\newblock In \emph{Proceedings of the IEEE/CVF conference on computer vision and pattern recognition}, 9308--9316.

\end{thebibliography}

\newpage
\appendix
\section*{\Large Appendix}

This Appendix includes detailed supplementary formulas and experimental results of our STDPose. Specifically, (1) we have provided a detailed derivation of the mutual information formulas and descriptions of the operations of several sub-modules. (2) Furthermore, we have supplemented additional experimental settings. (3) We have presented more ablation studies and more details of pseudo-label training. (4) Finally, we have displayed the visual results of our method on three benchmarks in challenging scenes.

\section{\large Supplementary Formulas}

In this section, we present some supplementary formulas regarding the model and mutual information objective to provide a detailed introduction to the proposed STDPose framework.

\textbf{Temporal Keypoints Synthesis (TKS) module.}\label{TKS_app}
The operations of the Temporal Keypoints Synthesis module can be expressed as:
\begin{equation}\label{eq_tks}
\begin{aligned}
\hat{H}_{t}^{i} &= Conv(H_{l}^{i,1} \oplus H_{t}^{i,1} \oplus H_{r}^{i,1}) , \cdots,\\
&\quad \oplus Conv(H_{l}^{i,j} \oplus H_{t}^{i,j} \oplus H_{r}^{i,j}) , \cdots, \\
&\quad \oplus Conv(H_{l}^{i,J} \oplus H_{t}^{i,J} \oplus H_{r}^{i,J}),\\
\tilde{H_{t}^{i}} &= Conv(Conv(\hat{H}_{t}^{i})  \oplus  H_{l}^{i}\oplus  H_{t}^{i}\oplus  H_{r}^{i}),
\end{aligned}
\end{equation}
where $\oplus$ is the concatenation operation and $Conv(\cdot)$ is the function of convolutional blocks. The superscript $j \in [1,2, \ldots, J]$ denotes $j$-th keypoint. $H_{t}^{i,j}$ represents the $j$-th keypoint heatmap of the key frame for person $i$. $\hat{H}_{t}^{i}$ represents the heatmaps after temporal merger.

\begin{table*}[t]
  \centering
  \renewcommand{\arraystretch}{1.1} 
  \begin{minipage}[t]{0.22\linewidth}
    \resizebox{\linewidth}{!}{
        \begin{tabular}{c|c|c}
        \hline
        $\theta$  &  $k$  & mAP \\
        \hline
         \multirow{5}{*}{$\theta = 0$}  & $k = 0.5$ & 90.6 \\
            & $k = 1$ & 90.6 \\
            & $k = 1.5$ & \textbf{90.7} \\
            & $k = 2$ & \textbf{90.7} \\
            & $k = 5$ & 90.6 \\     
        \hline
    \end{tabular}
    }
  \end{minipage}
  \begin{minipage}[t]{0.24\linewidth}
    \resizebox{\linewidth}{!}{
       \begin{tabular}{c|c|c}
        \hline
        $\theta$  &  $k$  & mAP \\
        \hline
         \multirow{5}{*}{$\theta = 0.2$}  & $k = 0.5$ & 90.6 \\
            & $k = 1$ & 90.7 \\
            & $k = 1.5$ & \textbf{90.8} \\
            & $k = 2$ & 90.7 \\
            & $k = 5$ & 90.7 \\     
        \hline
    \end{tabular}
    }
  \end{minipage}
  \begin{minipage}[t]{0.24\linewidth}
    \resizebox{\linewidth}{!}{
     \begin{tabular}{c|c|c}
        \hline
        $\theta$  &  $k$  & mAP \\
        \hline
         \multirow{5}{*}{$\theta = 0.5$}  & $k = 0.5$ & 90.7 \\
            & $k = 1$ & 90.8 \\
            & $k = 1.5$ & \textbf{90.9} \\
            & $k = 2$ & 90.8 \\
            & $k = 5$ & 90.8 \\     
        \hline
    \end{tabular}
    }
  \end{minipage}
  \begin{minipage}[t]{0.24\linewidth}
    \resizebox{\linewidth}{!}{
     \begin{tabular}{c|c|c}
        \hline
        $\theta$  &  $k$  & mAP \\
        \hline
         \multirow{5}{*}{$\theta = 0.7$}  & $k = 0.5$ & 90.6 \\
            & $k = 1$ & 90.7 \\
            & $k = 1.5$ & \textbf{90.8} \\
            & $k = 2$ & \textbf{90.8} \\
            & $k = 5$ & 90.7 \\     
        \hline
    \end{tabular}
    }
  \end{minipage}
  \captionof{table}{An ablation study was conducted on the parameters $k$ and $\theta$ of the modified sigmoid function. All results presented are from experiments performed on the PoseTrack2017 validation set.}  \label{table:sigmoid} 
\end{table*}

\textbf{Mutual Information (MI) objective.}\label{MIO_app} 
Mutual information measures the amount of information shared between random variables. Formally, the MI between two random variables \(x_{1}\) and \(x_{2}\) is defined as:
\begin{equation}\label{eq_mi1}
\begin{aligned}
\mathcal{I}\left(\boldsymbol{x}_{1} ; \boldsymbol{x}_{2}\right)=\mathbb{E}_{p\left(\boldsymbol{x}_{1}, \boldsymbol{x}_{2}\right)}\left[\log \frac{p\left(\boldsymbol{x}_{1}, \boldsymbol{x}_{2}\right)}{p\left(\boldsymbol{x}_{1}\right) p\left(\boldsymbol{x}_{2}\right)}\right],
\end{aligned}
\end{equation}
where \(p\left(\boldsymbol{x}_{1}, \boldsymbol{x}_{2}\right)\) is the joint probability distribution between $\boldsymbol{x}_{1}$ and $\boldsymbol{x}_{2}$, while $p\left(\boldsymbol{x}_{1}\right)$ and $p\left(\boldsymbol{x}_{2}\right)$ are their marginals. Within this framework, our main objective for learning effective temporal features and temporal poses can be formulated as:
\begin{equation}\label{eq_mi_max}
\begin{aligned}
\max \left[\mathcal{I}\left(y_{t}^{i} ;  \tilde{F_{t}^{i}} \mid F_{t}^{i}\right)+\mathcal{I}\left(y_{t}^{i} ; \tilde{H_{t}^{i}} \mid H_{t}^{i}\right)\right],
\end{aligned}
\end{equation}
where \(y_{t}^{i}\) denotes the pose label. The terms $\mathcal{I}\left(y_{t}^{i} ;  \tilde{F_{t}^{i}} \mid F_{t}^{i}\right)$ and $\mathcal{I}\left(y_{t}^{i} ; \tilde{H_{t}^{i}} \mid H_{t}^{i}\right)$ each represent the measure of task-relevant information contained within the fused feature \(\tilde{F_{t}^{i}}\) and the merged heatmaps \(\tilde{H_{t}^{i}}\), respectively, that is in addition to the information already present in $F_{t}^{i}$ and $H_{t}^{i}$. Optimizing this objective can maximize the task-relevant temporal information derived from the auxiliary frames. 

We utilize an streamlined computational approach due to the difficulty involved in computing the conditional mutual information. We initially simplify $\mathcal{I}\left(y_{t}^{i} ;  \tilde{F_{t}^{i}} \mid F_{t}^{i}\right)$ as follows:

\begin{equation}\label{eq_mi3}
\begin{aligned}
\mathcal{I}\left(y_{t}^{i} ;  \tilde{F_{t}^{i}} \mid F_{t}^{i}\right) =\mathcal{I}\left(y_{t}^{i} ; \tilde{F_{t}^{i}}\right) - &\mathcal{I}\left(\tilde{F_{t}^{i}} ; F_{t}^{i}\right) +\\
&\mathcal{I}\left(  \tilde{F_{t}^{i}} ;  F_{t}^{i} \mid  y_{t}^{i}\right),\\
\mathcal{I}\left(y_{t}^{i} ;  \tilde{F_{t}^{i}} \mid F_{t}^{i}\right) \rightarrow \mathcal{I}\left(y_{t}^{i} ; \tilde{F_{t}^{i}}\right) - &\mathcal{I}\left(\tilde{F_{t}^{i}} ; F_{t}^{i}\right),
\end{aligned}
\end{equation}
where \(\mathcal{I}\left(y_{t}^{i} ; \tilde{F_{t}^{i}}\right)\) denotes the relevance of the label $y_{t}^{i}$ and the fused feature $\tilde{F_{t}^{i}}$. $\mathcal{I}\left(\tilde{F_{t}^{i}} ; F_{t}^{i}\right)$ indicates the dependence between the fused feature $\tilde{F_{t}^{i}}$ and the key frame feature $F_{t}^{i}$. $\mathcal{I}\left(  \tilde{F_{t}^{i}} ;  F_{t}^{i} \mid  y_{t}^{i}\right)$ denotes the task-irrelevant information in both $\tilde{F_{t}^{i}}$ and $F_{t}^{i}$. 
We then approximate another term $\mathcal{I}\left(y_{t}^{i} ;  \tilde{H_{t}^{i}} \mid H_{t}^{i}\right)$ as follows:
\begin{equation}\label{eq_mi4}
\begin{aligned}
\mathcal{I}\left(y_{t}^{i} ;  \tilde{H_{t}^{i}} \mid H_{t}^{i}\right) = \mathcal{I}\left(y_{t}^{i} ; \tilde{H_{t}^{i}}\right) - &\mathcal{I}\left(\tilde{H_{t}^{i}} ; H_{t}^{i}\right) +\\
&\mathcal{I}\left(  \tilde{H_{t}^{i}} ;  H_{t}^{i} \mid  y_{t}^{i}\right),\\
\mathcal{I}\left(y_{t}^{i} ; \tilde{H_{t}^{i}} \mid H_{t}^{i}\right) \rightarrow \mathcal{I}\left(y_{t}^{i} ; \tilde{H_{t}^{i}}\right) - &\mathcal{I}\left(\tilde{H_{t}^{i}} ; H_{t}^{i}\right),
\end{aligned}
\end{equation}
where $\mathcal{I}\left(  \tilde{H_{t}^{i}} ;  H_{t}^{i} \mid  y_{t}^{i}\right)$ represents the task-irrelevant information in both $\tilde{H_{t}^{i}}$ and $H_{t}^{i}$. \(\mathcal{I}\left(y_{t}^{i} ; \tilde{H_{t}^{i}}\right)\) denotes the relevance of the label $y_{t}^{i}$ and the merged heatmaps $\tilde{H_{t}^{i}}$. $\mathcal{I}\left(\tilde{H_{t}^{i}} ; H_{t}^{i}\right)$ indicates the dependence between the merged pose heatmaps $\tilde{H_{t}^{i}}$ and the key frame pose heatmaps $H_{t}^{i}$. Optimizing the task objective heuristically tends to give significant prominence to the task-specific information, overshadowing the task-irrelevant information. Given sufficient training, we can reasonably posit that the influence of task-irrelevant information will diminish to negligible levels, and during optimization, it is appropriate to remove $\mathcal{I}\left(  \tilde{F_{t}^{i}} ;  F_{t}^{i} \mid  y_{t}^{i}\right)$ and $\mathcal{I}\left(  \tilde{H_{t}^{i}} ;  H_{t}^{i} \mid  y_{t}^{i}\right)$.

Through the above simplifications, the MI loss $\mathcal{L}_{\mathrm{MI}}$ can be more specifically expressed as:
\begin{equation}\label{eq_mi_add}
\begin{aligned}
\mathcal{L}_{\mathrm{MI}}  = -&\alpha \cdot \left[\mathcal{I}\left(y_{t}^{i} ; \tilde{F_{t}^{i}}\right) - \mathcal{I}\left(\tilde{F_{t}^{i}} ; F_{t}^{i}\right)\right] - \\
&\beta \cdot \left[\mathcal{I}\left(y_{t}^{i} ; \tilde{H_{t}^{i}}\right) - \mathcal{I}\left(\tilde{H_{t}^{i}} ; H_{t}^{i}\right)\right].
\end{aligned}
\end{equation}

\textbf{Cross-attention layer.}\label{CA} 
Our Pose Aggregation block's Cross-attention layer takes two inputs: the output $Z$ from the previous Self-attention layer and the fused feature $\tilde{F_{t}^{i}}$. We perform a linear mapping on $Z$ to transform it into the query $Q$, and map $\tilde{F_{t}^{i}}$ into the key $K$ and the value $V$. Then, we carry out the attention computation on them. These operations can be formalized as follows:
\begin{equation}
\label{eq_cross_atten}
\begin{aligned}
Q &= Z \otimes  W_{Q},\\
K &= \tilde{F_{t}^{i}} \otimes  W_{K},\\
V &= \tilde{F_{t}^{i}} \otimes  W_{V},\\
Atten(Q, K, V) &= Softmax\left(\frac{Q K^{T}}{\sqrt{D}}\right) V,
\end{aligned}
\end{equation}
where $\otimes $ represents matrix multiplication. $W_{Q}$, $W_{K}$, and $W_{V}$ are three learnable matrices responsible for mapping the inputs. $D$ is the value of the embedding dimension. $K^{T}$ is the transpose of the matrix $K$. $Atten(\cdot)$ denotes the attention computation, and $Softmax(\cdot)$ refers to the softmax calculation.

\textbf{SpatioTemporal Dynamics Aggregation (STDA) module.}\label{app_STDA}
The operations of the SpatioTemporal Dynamics Aggregation module can be expressed as:
\begin{equation}\label{eq11}
\begin{aligned}
\hat{H_{t}^{i}} &= E_{patch}((\tilde{H_{t}^{i}} \odot  M_{t}^{i} ) \oplus \tilde{H_{t}^{i}}) + E_{pos} ,\\
\hat{H_{t}^{N}} &= PA_{N}(,\cdots, (PA_{n}(,\cdots,(PA_{1}(\hat{H_{t}^{i}}, \tilde{F_{t}^{i}}))))) , \\
\bar{H_{t}^{i}} &= Head(\hat{H_{t}^{N}}),
\end{aligned}
\end{equation}
where $E_{patch}(\cdot)$ and $E_{pos}$ respectively represent the patch embedding layer and position embedding. $\odot$ denotes the dot product operation. The subscript $n \in [1,2, \ldots, N]$ of $PA_{n}(\cdot)$ indicates the $n$-th Pose Aggregation block. $\hat{H_{t}^{N}}$ is the output of the last Pose Aggregation block. $Head(\cdot)$ denotes a classical pose detection head, which incorporates two layers dedicated to the upsampling process.

\begin{table*}[t]
    \centering
    \renewcommand{\arraystretch}{1.2} 
    \resizebox{0.98\textwidth}{!}{
    \begin{tabular}{c|c|c|ccccccc|c}
        \hline
         Model & Backbone & Image Size & Head & Shoulder & Elbow & Wrist & Hip & Knee & Ankle & Mean \\
        \hline
         DCPose~\cite{liu2021deep} & HRNet~\cite{sun2019deep} & 384$\times$288 & 91.2 & 90.8 & 88.4 & 84.3 & 87.7 & 86.7 & 83.2 & 87.7 \\
         \hline
         \multirow{3}{*}{STDPose (Ours)} & \multirow{3}{*}{ViT~\cite{dosovitskiy2020image}} & 224$\times$168 &   92.4 & 92.8 & 90.6 & 87.7 & 91.0 & 89.9 & 88.6 & 90.6 \\
          & &  256$\times$192  & 92.7 & 93.1 & 91.4 & 88.1 & 91.9 & 90.3 & 88.1 & 90.9  \\  
          & &  384$\times$288  & \textbf{93.3} & \textbf{93.4} & \textbf{92.1} & \textbf{89.3} & \textbf{92.7} & \textbf{91.2} & \textbf{89.2} & \textbf{91.7} \\         
        \hline
    \end{tabular}
    }
    \caption{Performance evaluation of STDPose with various input resolutions and comparison with DCPose~\cite{liu2021deep} on the PoseTrack2017 validation set.}\label{table:image_size}
\end{table*}

\section{\large Additional Experimental Settings}\label{app_settings}

\textbf{Datasets.} PoseTrack, a comprehensive benchmark suite for video-based human pose estimation and tracking, encompasses a diverse array of challenges such as crowded scenes and rapid movements. The \textbf{PoseTrack2017} dataset~\cite{iqbal2017posetrack}, adhering to the official protocol, is composed of 250 video sequences for training, 50 for validation, and an additional 214 for testing, amassing a total of 80,144 pose annotations. Each of these sequences is meticulously annotated with 15 key points, augmented by a visibility flag indicating the state of each joint. Expanding on its predecessor, \textbf{PoseTrack2018}~\cite{andriluka2018posetrack} introduces 1,138 video sequences with a notable rise to 153,615 annotations, divided into 593 for training, 170 for validation, and 375 for testing. Each individual is meticulously annotated with 15 joints and an added visibility flag. \textbf{PoseTrack2021}~\cite{doering2022posetrack21}, the latest installment, not only extends these annotations to an impressive 177,164 but also focuses on challenging scenarios like small persons and dense crowds. A key enhancement in this iteration is the refined joint visibility flag, which improves the dataset's handling of occlusions and enriches its real-world applicability.

\textbf{Implementation details.} Our STDPose framework is implemented by PyTorch 1.9. We incorporate data augmentation including random rotation [$-45^\circ$, $45^\circ$], random scale [0.65, 1.35], truncation (half body), and flipping during training. We adopt the AdamW optimizer with a base learning rate of 2e\text{-}4 (decays to 2e\text{-}5 and 2e\text{-}6 at the $12^{th}$ and $16^{th}$ epochs, respectively). We train the model using 4 Nvidia Geforce RTX 2080 Ti GPUs. All training process is terminated within 20 epochs.

\begin{figure*}[t]
\centering
\includegraphics[width=.88\linewidth]{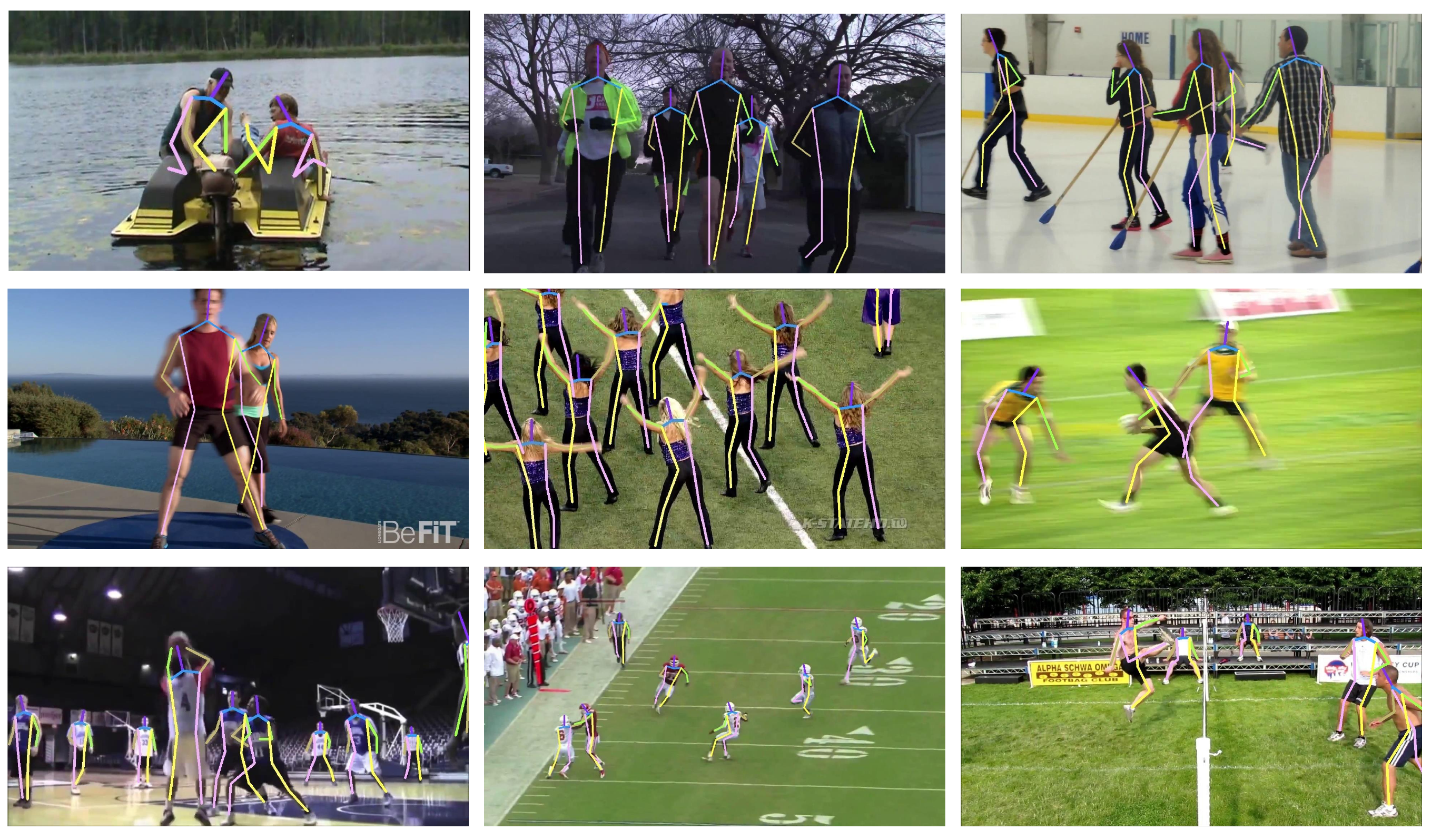}
\caption{Visual results of our STDPose on the PoseTrack2017~\cite{iqbal2017posetrack} dataset include challenging scenes, such as rapid movements and pose occlusions.}
\label{fig:vis_17}
\end{figure*}

\section{\large Additional Experimental Results}\label{app_results}
In this section, we first examine the feasibility of training pose estimation using pseudo-labels generated through pose propagation. Then, we investigate the influence of the parameters of the modified sigmoid function within the Pose Decoder. We further explore the impact of input images of different sizes on the results.

\textbf{Pseudo-label training for pose estimation.} 
To obtain pseudo-labels, we first train our STDPose on sparsely-labeled videos from the training set of PoseTrack2017. When $T$ is set to 2, it means that labeled and unlabeled frames each account for half of the video. Increasing $T$ will reduce the proportion of labeled frames. Note that only 30 consecutive frames in the middle of each video are labeled. We utilize the well-trained pose propagation model to perform the pose propagation task to generate pseudo-ground truth labels for unlabeled frames. Subsequently, we train the pose estimation model using these pseudo-labels along with the labels from labeled frames. Finally, we evaluate the results of pose estimation on the validation set of PoseTrack2017.

All results are presented in Table~\ref{table:pseudo}, from which we can draw several conclusions. We observe that the lower the value of $T$ (\textit{i.e.}, the higher the proportion of labeled frames), the better the results of pose estimation. 
Firstly, when $T$ is set to 7, only 16.7\% of the labeled frames are used for training, and there is a noticeable decrease in performance. This suggests that there is not enough useful data to learn pose detection in this scenario. 
Secondly, we observe a significant enhancement in the performance of our model, notably improving to 84.3 mAP, which is close to that of FAMI-Pose~\cite{liu2022temporal}, when $T$ is reduced to 4. 
We further decreased $T$ to 2 and once again achieved a performance boost, with our model surpassing FAMI-Pose and reaching 85.2 mAP. This convincing result demonstrates that our approach can achieve comparable results to previous state-of-the-art methods with only half of the labeled data. 
Additionally, we noticed that when 
$T$ is set to 4, our model's detection accuracy for the ankle joint exceeded FAMI-Pose by 2.0 mAP, which again proves the strong robustness of our model. 
Overall, these results of our method present a fact that high pose estimation accuracy can be achieved with only a small portion of labeled data, thereby truly reducing the dependence on large-scale datasets with dense annotations.

\textbf{Study on parameters of the modified sigmoid function.} 
We conducted extensive experiments to determine the optimal values for the parameters $k$ and $\theta$ of the modified sigmoid function, as detailed in Table~\ref{table:sigmoid}. It is evident from the ablation study that incorporating the modified sigmoid function has significantly improved the performance of our framework. The results in Table~\ref{table:sigmoid} clearly demonstrate that adjusting these parameters can lead to performance fluctuations, but the variations are not significant. We speculate a possible reason for this is the automatic filtering out of non-essential areas by the mask obtained after the softmax operation. This occurs when the mask is multiplied with $\tilde{H_{t}^{i}}$ during the STDA process, as all areas except those around the keypoints in $\tilde{H_{t}^{i}}$ contain extremely low values.

\textbf{Study on different input sizes} Previous methods~\cite{liu2021deep, feng2023mutual,liu2022temporal} predominantly utilize HRNet~\cite{sun2019deep} as the backbone network, with input image sizes typically being 384$\times$288. However, We adopt Vision Transformer~\cite{dosovitskiy2020image} as the backbone network, which commonly takes images of size 256$\times$192 as input. Intuitively, we speculate that increasing the input image size will lead to some performance improvement, while reducing the size will decrease computation but also lead to a decline in performance. To evaluate the impact of different input resolutions on our STDPose, we train STDPose with various input image sizes and present the results in Table~\ref{table:image_size}. We observe that the outcomes are largely consistent with our speculations. When we reduce the input image size to 224$\times$168, the performance of our model slightly deteriorates. Conversely, when we increase the size to 384$\times$288, there is a notable enhancement in performance, achieving 91.7 mAP.

\begin{figure*}[t]
\centering
\includegraphics[width=.88\linewidth]{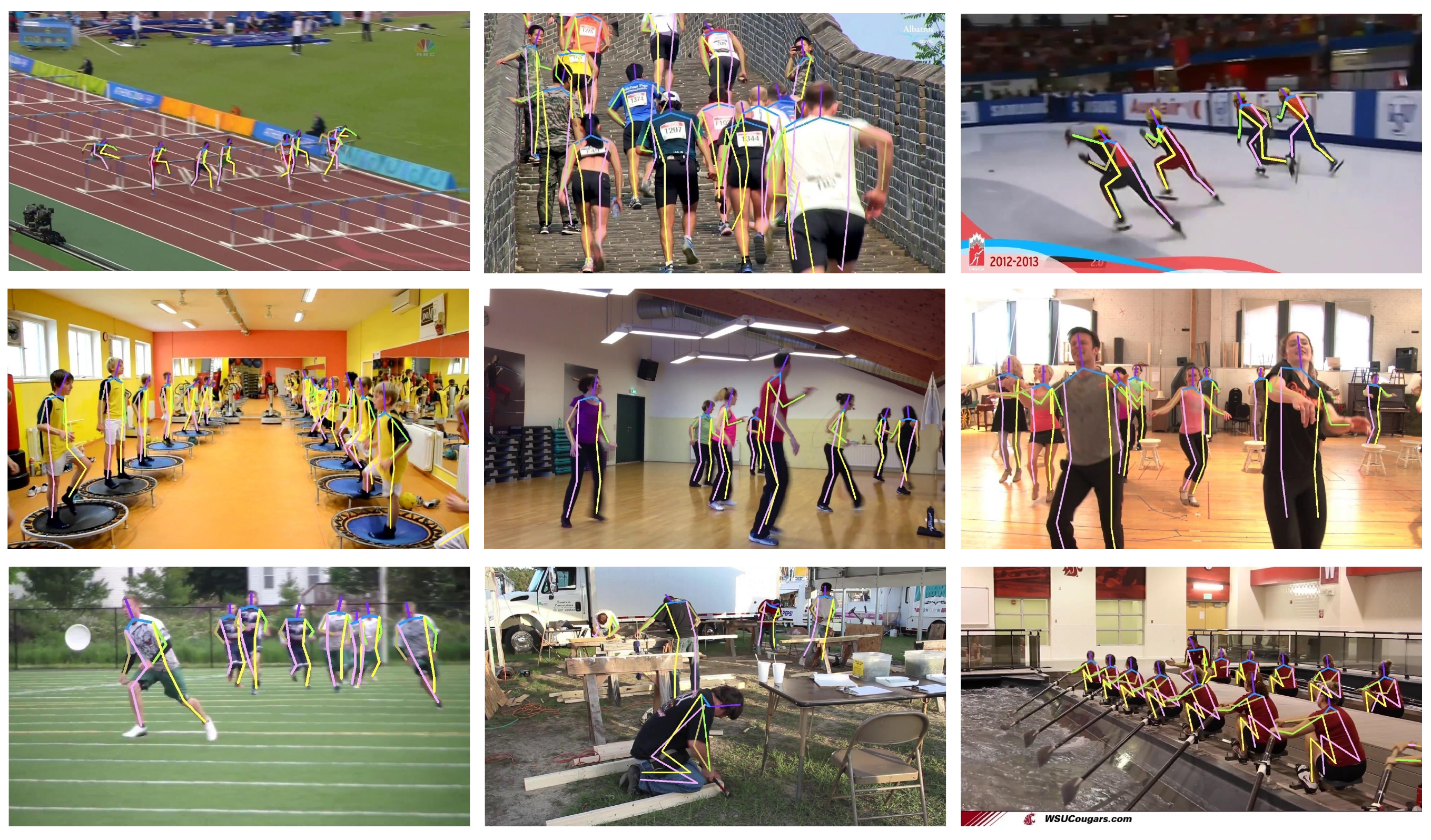}
\caption{Visual results of our STDPose on the PoseTrack2018~\cite{andriluka2018posetrack} dataset include challenging scenes, such as rapid movements and pose occlusions.}
\label{fig:vis_18}
\end{figure*}

\begin{figure*}[t]
\centering
\includegraphics[width=.88\linewidth]{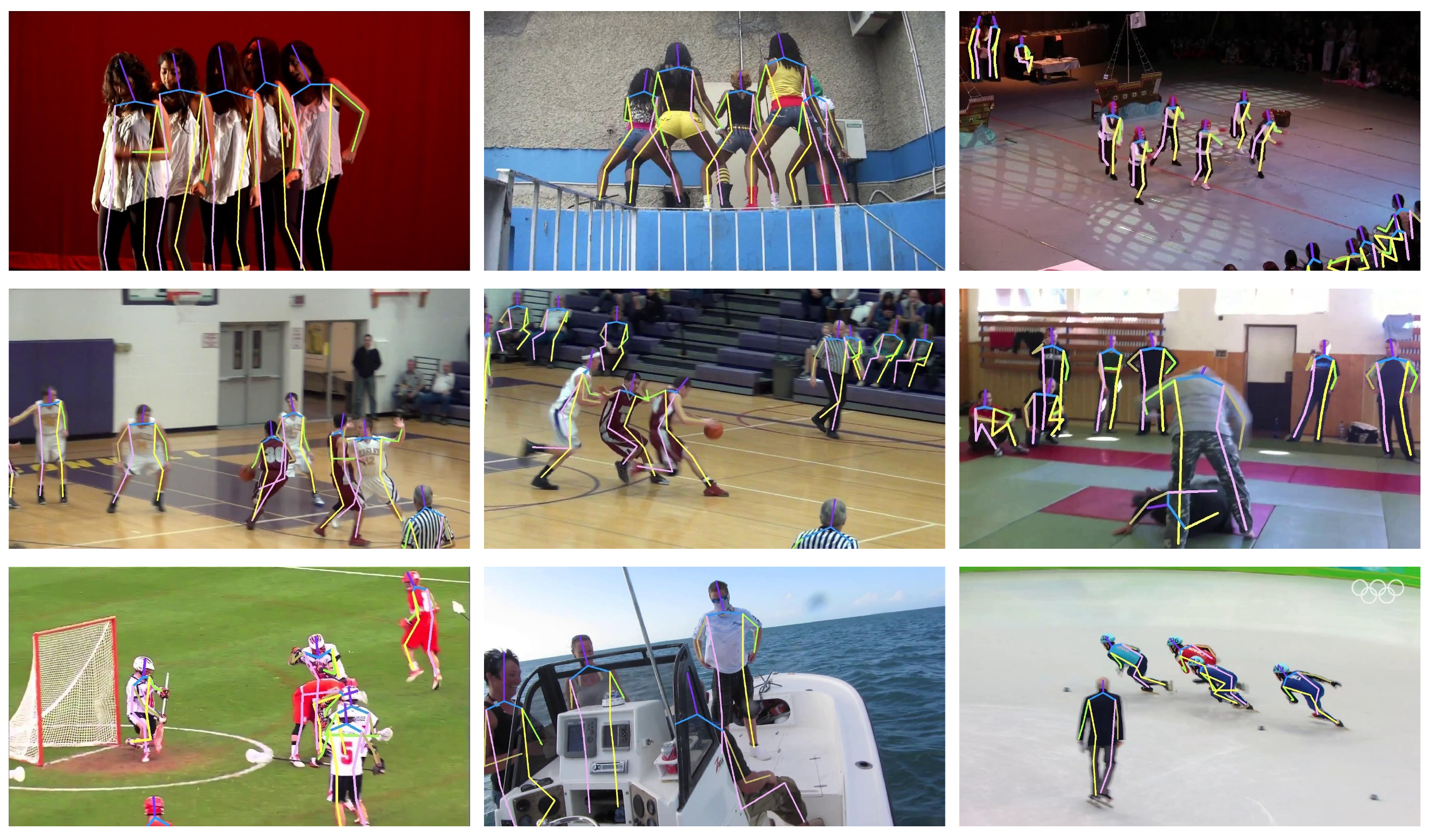}
\caption{Visual results of our STDPose on the PoseTrack2021~\cite{doering2022posetrack21} dataset include challenging scenes, such as rapid movements and pose occlusions.}
\label{fig:vis_21}
\end{figure*}

\section{\large Visualized Results in Challenging Scenarios}\label{app_vis}

In this section, we present visualized results for scenarios featuring intricate spatiotemporal dynamics, such as occlusion and blur, on the PoseTrack2017~\cite{iqbal2017posetrack}, PoseTrack2018~\cite{andriluka2018posetrack}, and PoseTrack21~\cite{doering2022posetrack21} datasets, as depicted in Figures~\ref{fig:vis_17}, \ref{fig:vis_18}, and ~\ref{fig:vis_21}. These results substantiate the robustness of our proposed approach.

\end{document}